\documentclass[a4paper]{article}
\usepackage{amsmath,amssymb,amsthm,url}

\usepackage{algorithm,algorithmic}
\makeatletter
  \renewcommand{\ALG@name}{Protocol}
\makeatother

\newcommand{\e}{\mathrm{e}}
\newcommand{\EXP}{\mathrm{E}}
\newcommand{\PPP}{\mathcal{P}}
\newcommand{\Omin}{\mathcal{M}}

\newtheorem{theorem}{Theorem}
\newtheorem{lemma}[theorem]{Lemma}
\renewcommand{\proofname}{Proof}
\theoremstyle{definition}
\newtheorem{remark}[theorem]{Remark}
\newtheorem{assumption}{Assumption}

\begin{document}

\title{Supermartingales in Prediction with Expert Advice}

\author{Alexey Chernov,
Yuri Kalnishkan,
Fedor Zhdanov,\\
Vladimir Vovk\\[1ex]
\normalsize  Computer Learning Research Centre\\
\normalsize  Department of Computer Science\\
\normalsize  Royal Holloway, University of London,\\
\normalsize  Egham, Surrey TW20 0EX, UK\\
\normalsize\texttt{\{chernov, yura, fedor, vovk\}@cs.rhul.ac.uk}
}

\maketitle

\begin{abstract}
We apply the method of defensive forecasting,
based on the use of game-theoretic supermartingales, 
to prediction with expert advice.
In the traditional setting 
of a countable number of experts and a finite number of outcomes,
the Defensive Forecasting Algorithm
is very close to the well-known Aggregating Algorithm.
Not only the performance guarantees
but also the predictions are the same 
for these two methods
of fundamentally different nature.
We discuss also a new setting
where the experts can give advice
conditional on the learner's future decision.
Both the algorithms can be adapted to the new setting
and give the same performance guarantees 
as in the traditional setting.
Finally, we outline
an application of defensive forecasting to a setting
with several loss functions.
\end{abstract}

\section{Introduction}

The framework of prediction with expert advice 
was introduced in the late 1980s.
In contrast to statistical learning theory,
the methods of prediction with 
expert advice do not require 
statistical assumptions about the source of data.
The role of the assumptions is played by 
a ``pool of experts'': 
the forecaster, called Learner, bases his predictions upon
the predictions and performance of the experts.
For details and references, see the monograph~\cite{CBL:2006}.

Many methods for prediction with expert advice are known.
This paper deals with two of them: 
the Aggregating Algorithm~\cite{Vovk:1990}
and defensive forecasting~\cite{Vovk:2005}.
The Aggregating Algorithm (the AA for short)
is a member of the family of exponential-weights algorithms
and implements a Bayesian-type aggregation;
various optimality properties of the AA have been 
established~\cite{Vovk:1998}.
Defensive forecasting is a recently developed technique
that combines the ideas of 
ga\-me-the\-o\-re\-tic probability~\cite{ShV:2001}
with Levin and G\'acs's ideas
of neutral measure~\cite{gacs:2005,levin:1976uniform}
and Foster and Vohra's ideas of universal calibration~\cite{FV:1998}.

The idea of defensive forecasting
comes from an interpretation of probability
with the help of perfect information games.
The Learner develops his strategy
modeling a game where 
a probability forecaster plays 
on the actual data
against an imaginary opponent, Sceptic,
that represents a law of probability.
The capital of Sceptic tends to infinity (or becomes large) 
if the players' moves lead to violation of this law.
The capital of a strategy for Sceptic 
as a function of other players' moves
is called a (ga\-me-the\-o\-re\-tic) supermartingale.
It is known
(see Lemma~\ref{lem:LevinSuper} in this paper)
that for any supermartingale there is a forecasting strategy
that prevents this supermartingale from growing 
(``defending'' against this strategy of Sceptic),
thereby forcing the corresponding law of probability.
The older versions of defensive forecasting
(see, e.g., \cite{Vovk:2005}) minimize Learner's actual loss 
with the help of the following trick: 
a forecasting strategy
is constructed so that the actual losses
(Learner's and experts') are close to the 
(one-step-ahead conditional) expected losses;
at each step Learner minimizes the expected loss
(that is, the law of probability used in this case 
is the conjunction of several laws of large numbers).
This paper gives a self-contained description
of a different version of the defensive forecasting method.
We use certain supermartingales and 
do not need to talk about the underlying laws of probability.

Defensive forecasting, as well as the AA, can be used for 
competitive online prediction against ``pools of experts'' 
consisting of all functions from a large function class
(see~\cite{Vovk:2006,Vovk:2006m}). 
However, the loss bounds proved so far are generally incomparable:
for large classes (such as many Sobolev spaces),
defensive forecasting is better, whereas for smaller classes
(such as classes of analytical functions), 
the AA works better.
Note that the optimality results for the AA are obtained 
for experts that are free agents, 
not functions from a given class;
thus we need to evaluate the algorithms anew.
This general task requires a deeper understanding
of the properties of defensive forecasting.

In this paper, the AA and defensive forecasting are discussed
in the simple case of a finite number of outcomes.
Learner competes with a countable pool of Experts $\Theta$. 
Experts and Learner give predictions and 
suffer some loss at each step.
A game is a specification what predictions are admissible
and what losses a prediction incur for each outcome.
For every game, 
we are interested in performance guarantees of the form
$$
\forall\theta\in\Theta\:\forall N\quad L_N\le cL_N^\theta + a^\theta\,,
$$
where $L_N$ is the cumulative loss of Learner and
$L_N^\theta$ is the cumulative loss of Expert~$\theta$
over the first $N$ steps,
$c$ is some constant and $a^\theta$ depends on $\theta$ only.
Section~\ref{sec:AA} recalls the AA and its loss bound
(Theorem~\ref{thm:AAVovk})
and introduces notation used in the paper.

Section~\ref{sec:super} presents the main results of the paper.
Subsection~\ref{ssec:df} 
describes the Defensive Forecasting Algorithm (DFA),
which is based on the use of ga\-me-the\-o\-re\-tic supermartingales,
and its loss bound (Theorem~\ref{thm:superbound}).
It turns out that if the AA and the DFA
are both applicable to a game,
they guarantee the same loss bound.
Subsections~\ref{ssec:DFtoAA}--\ref{ssec:constrproper}
discuss when the DFA and the AA are applicable.
Loosely speaking,
if the DFA is applicable then the AA is applicable as well
(Theorem~\ref{thm:superAA});
and for games satisfying some additional assumptions,
if the AA is applicable then the DFA is applicable 
(Theorems~\ref{thm:superconstr} and Theorem~\ref{thm:proper}).
Subsection~\ref{ssec:DFrevisited} 
gives a criterion of the AA realizability
in terms of supermartingales (Theorem~\ref{thm:criterion})
using a rather awkward variant of the DFA.
The construction of the supermartingales used in this paper
involves a parameterization of the game
with the help of a proper loss function.
Proper loss functions play an important role 
in Bayesian statistics,
and their meaning in our context 
is discussed in Subsections~\ref{ssec:proper}
and~\ref{ssec:constrproper}.

The rest of the paper is devoted to 
modifications of the standard setting. 
Subsection~\ref{ssec:cont} applies the DFA
in an extended setting where the outcomes form 
a finite-dimensional simplex.
Section~\ref{sec:second-guess}
introduces a new setting for prediction
with expert advice,
where the experts are allowed to ``second-guess'',
that is, to give 
``conditional'' predictions that are functions
of the future Learner's decision
(cf.~the notion of internal regret~\cite{FV:1999}).
If the dependence is regular enough (namely, continuous),
the DFA works in the new setting virtually without changes
(Theorem~\ref{thm:supersecond}).
The AA with some modification based on the fixed point theorem
can be applied in the new setting too
(Theorem~\ref{thm:AAfixed}).
Section~\ref{sec:multloss} briefly outlines one more application
of the DFA: a setting with several loss functions.

Some results of the paper appeared in~\cite{Vovk:2007}
and in ALT'08 proceedings~\cite{CKZV:2008}.

\section{Games of Prediction and the Aggregating Algorithm}
\label{sec:AA}

We begin with formulating the setting of prediction
with expert advice. 
A \emph{game of prediction} consists of three components:
a non-empty set $\Omega$ of possible outcomes,
a non-empty set $\Gamma$ of possible decisions,
and a function
${\lambda\colon\Gamma\times\Omega\to[0,\infty]}$
called the \emph{loss function}.
In this paper we assume that the set $\Omega$ is finite.

The set
$\Lambda=
\{\,g\in[0,\infty]^\Omega\mid\exists\gamma\in\Gamma\,\forall\omega\in\Omega\:
g(\omega)=\lambda(\gamma,\omega)\,\}$ 
is called the set of \emph{predictions}
of the game.
In this paper, we will identify each decision $\gamma\in\Gamma$
with the function $\omega\mapsto \lambda(\gamma,\omega)$ 
(and also with a point in a 
$\lvert\Omega\rvert$-dimensional Euclidean space
with pointwise operations).
A loss function can be considered as a parameterization of $\Lambda$
by elements of $\Gamma$.
To study the properties of a game,
we do not need to know the decision set $\Gamma$
and the loss function;
we can forget about them and consider the prediction set $\Lambda$
only.
\emph{From now on, a game will by specified by a pair} 
$(\Omega,\Lambda)$, where $\Lambda\subseteq[0,\infty]^\Omega$.
We will use the letter $\gamma$ (as well as $g$)
with indices to denote elements of $[0,\infty]^\Omega$
(rather than decisions).

However, loss functions remain a convenient method to specify
a game, and we will use them in examples.
Also an important technical tool will be 
a kind of canonical parameterization of $\Lambda$
given by the so called proper loss functions.
Also loss functions are unavoidable in Section~\ref{sec:multloss},
where we consider games with several simultaneous losses.

The game of prediction with expert advice is played 
by Learner, Experts, and Reality;
the set (``pool'') of Experts is denoted by~$\Theta$.
We will assume that $\Theta$ is (finite or) countable.
There is no loss of generality in assuming
that Reality and all Experts are cooperative,
since we are only interested in what can be achieved by Learner alone;
therefore, we essentially consider a two-player game.
The game is played according to Protocol~\ref{prot:PEA}.

\begin{algorithm}[ht]
  \caption{Prediction with Expert Advice}
  \label{prot:PEA}
  \begin{algorithmic}
    \STATE $L_0:=0$.
    \STATE $L_0^\theta:=0$, for all $\theta\in\Theta$.
    \FOR{$n=1,2,\dots$}
      \STATE All Experts $\theta\in\Theta$ announce $\gamma_n^\theta\in\Lambda$.
      \STATE Learner announces $\gamma_n\in\Lambda$.
      \STATE Reality announces $\omega_n\in\Omega$.
      \STATE $L_n:=L_{n-1}+\gamma_n(\omega_n)$.
      \STATE $L_n^\theta:=L_{n-1}^\theta+\gamma_n^\theta(\omega_n)$, 
                        for all $\theta\in\Theta$.
    \ENDFOR
  \end{algorithmic}
\end{algorithm}

The goal of Learner is 
to keep $L_n$ smaller or at least not much greater than $L_n^\theta$, 
at each step $n$ and for all $\theta\in\Theta$.

To analyze the game,
we need some additional notation.
A point $g\in[0,\infty]^\Omega$ is called 
a \emph{superprediction} in the game $(\Omega,\Lambda)$
if there is $\gamma\in\Lambda$
such that 
${\gamma(\omega)\le g(\omega)}$ for all $\omega\in\Omega$.
It is convenient to write the last condition as
$\gamma\le g$.
In the sequel, we will use pointwise relations and operations
for the elements of $[0,\infty]^\Omega$
without special mentioning.

For a game $(\Omega,\Lambda)$,
denote by $\Sigma_\Lambda$ the set of all superpredictions.
Using operations on sets, this definition can be written as
$\Sigma_\Lambda=\Lambda+[0,\infty]^\Omega=
{\{\gamma+g\mid \gamma\in\Lambda,\;g\in[0,\infty]^\Omega\}}
$.

The \emph{Aggregating Algorithm} is a strategy for Learner.
It has four parameters: reals $c\ge 1$ and $\eta>0$,
a \emph{distribution} $P_0$ on $\Theta$
(that is,
$P_0(\theta)\in[0,1]$ for every $\theta\in\Theta$ 
and $\sum_{\theta\in\Theta} P_0(\theta)=1$),
and a \emph{substitution function} 
$\sigma\colon\Sigma_\Lambda\to\Lambda$
such that $\sigma(g)\le g$
for any $g\in\Sigma_\Lambda$. 

At step $N$, the AA computes 
$g_N\in[0,\infty]^\Omega$ by the formula 
$$
g_N(\omega) =
 -\frac{c}{\eta}\ln\left(\sum_{\theta\in\Theta}
\frac{P_{N-1}(\theta)}{\sum_{\theta\in\Theta}P_{N-1}(\theta)}
\exp(-\eta\gamma_N^\theta(\omega))\right)\,,
$$
where
$$
P_{N-1}(\theta)=P_0(\theta)
\prod_{n=1}^{N-1}\exp(-\eta\gamma_n^\theta(\omega_n))
$$
is the (posterior) distribution on $\Theta$.
Then, $\gamma_N=\sigma(g_N)$ is announced as 
Learner's prediction.

The step~$N$ of the AA can be performed 
if and only if $g_N$ is a superprediction
($g_N\in\Sigma_\Lambda$),
that is, if
\begin{equation}\label{eq:AA}
\exists \gamma_N\in\Lambda\;\forall\omega\quad
\gamma_N(\omega)\le 
 -\frac{c}{\eta}\ln\left(\sum_{\theta\in\Theta}
\frac{P_{N-1}(\theta)}{\sum_{\theta'\in\Theta}P_{N-1}(\theta')}
\exp(-\eta\gamma_N^\theta(\omega))\right)\,.
\end{equation}

We say that the AA is $(c,\eta)$-\emph{realizable}
(for the game $(\Omega,\Lambda)$)
if condition~\eqref{eq:AA} is true regardless of 
$\Theta$, $N$,
$\gamma_N^\theta\in\Lambda$ and $P_{N-1}$
(that is, regardless of $P_0$, the history of the previous moves, 
and the opponents' moves at the last step).
This requirement can be restated in several
equivalent forms:
for any finite set $G\subseteq\Lambda$ and
for any distribution $\rho$ on $G$, it holds that
\begin{equation}\label{eq:realizLambda}
\exists \gamma\in\Lambda\quad
\gamma\le
-\frac{c}{\eta}\ln\left(\sum_{g\in G}\rho(g)\exp(-\eta g)\right);
\end{equation}
or equivalently, 
for any finite $G\subseteq\Sigma_\Lambda$ and
any distribution $\rho$ on $G$, it holds that
\begin{equation}\label{eq:realiz}
\exists \gamma\in\Sigma_\Lambda\quad
\gamma\le 
-\frac{c}{\eta}\ln\left(\sum_{g\in G}\rho(g)\exp(-\eta g)\right);
\end{equation}
equivalently, in the last formula $\le$ can be replaced by~$=$.
Indeed, 
the condition~\eqref{eq:AA} implies~\eqref{eq:realizLambda}
since $\gamma_N^\theta$ and $P_0$ are arbitrary;
$G\subseteq\Lambda$ 
can be replaced by $G\subseteq\Sigma_\Lambda$
since the right-hand side of~\eqref{eq:realizLambda}
increases when elements of $G$ increase;
by definition, \eqref{eq:realizLambda} means
that its right-hand side belongs to $\Sigma_\Lambda$,
and we get~\eqref{eq:realiz} with~$=$ instead of~$\le$.
Clearly, \eqref{eq:AA} follows from~\eqref{eq:realiz},
if we allow countably infinite $G$ as well
(then we can take $\{\gamma_N^\theta\mid \theta\in\Theta\}$ for $G$),
which is possible due to the following property
of convex sets.

For a given $\eta$,
the \emph{exp-convex hull} of $\Sigma_\Lambda$
is the set $\Sigma_\Lambda^\eta\supseteq\Sigma_\Lambda$ that consists
of all points in $[0,\infty]^\Omega$ of the form 
\begin{equation}\label{eq:expSigma}
\log_{(\e^{-\eta})}\left(\sum_{g\in G}\rho(g)
   \bigl(\e^{-\eta}\bigr)^{g}\right)
=-\frac{1}{\eta}\ln\left(\sum_{g\in G}\rho(g)
   \exp(-\eta g)\right)\,,
\end{equation}
where $G$ is a finite subset of $\Sigma_\Lambda$ and
$\rho$ is a distribution on $G$.
Actually, $\exp(-\eta\Sigma_\Lambda^\eta)$
is the convex hull of $\exp(-\eta\Sigma_\Lambda)$.
As known from convex analysis,
we get the same definition if we allow infinite $G$ 
(see e.\,g.~\cite[Theorem~2.4.1]{Blackwell:1954}).
With this notation,
the condition~\eqref{eq:realiz} says that
$\Sigma_\Lambda\supseteq c\Sigma_\Lambda^\eta$.

Let us state some properties of the set $\Sigma_\Lambda^\eta$.
First,
$\Sigma_\Lambda^\eta=\Sigma_\Lambda^\eta+[0,\infty]^\Omega$,
that is,
if $\Sigma_\Lambda^\eta$ is a prediction set
then its superprediction set is $\Sigma_\Lambda^\eta$ itself.
(Indeed, if a point $g_0$ of the form~\eqref{eq:expSigma} 
belongs to $\Sigma_\Lambda^\eta$
as a combination of $g_i\in G\subseteq\Sigma_\Lambda$
then, for any $g\in[0,\infty]^\Omega$,
the point $g_0+g$ belongs to $\Sigma_\Lambda^\eta$ 
as the combination of $g_i+g$.)
The set $\exp(-\eta\Sigma_\Lambda^\eta)$ is convex
(clearly, the points of the form~\eqref{eq:expSigma} belong
to $\Sigma_\Lambda^\eta$
also if we allow $G\subseteq\Sigma_\Lambda^\eta$).
The convexity of exponent implies that 
the set $\Sigma_\Lambda^\eta$ is convex as well
(if $g_1,g_2\in\Sigma_\Lambda^\eta$ then 
$\alpha g_1+(1-\alpha)g_2
\ge 
-{\frac{1}{\eta}\ln\bigl(
       \alpha\exp(-\eta g_1)+(1-\alpha)\exp(-\eta g_2)
      \bigr)}$
and hence $\alpha g_1+(1-\alpha)g_2\in\Sigma_\Lambda^\eta$ too).

The game $(\Omega,\Lambda)$ is called \emph{$\eta$-mixable}
if the AA is $(1,\eta)$-realizable, that is,
if $\Sigma_\Lambda = \Sigma_\Lambda^\eta$.
The game is \emph{mixable} if it is 
$\eta$-mixable for some $\eta>0$.
The mixable games are of special interest.
In a sense,
the AA works with mixable games only,
and 
to any non-mixable game $(\Omega,\Lambda)$ the AA assigns
the $\eta$-mixable game $(\Omega,\Sigma_\Lambda^\eta)$
and then simply transfers
the loss bound (at the price of a constant factor).
Standard examples of mixable games are
the square loss game~\cite[Example~4]{Vovk:1998},
which is $\eta$-mixable for $\eta\in(0,2]$,
and the logarithmic loss game~\cite[Example~5]{Vovk:1998},
which is $\eta$-mixable for $\eta\in(0,1]$;
see Subsection~\ref{ssec:examples}.
A standard example of a non-mixable game
is the absolute loss game~\cite[Example~3]{Vovk:1998}
with the loss function
$\lambda(p,\omega)=\lvert p-\omega\rvert$, $p\in[0,1]$,
$\omega\in\{0,1\}$
(its prediction set $\Lambda$ is 
$\{(x,y)\in[0,1]^2\mid x+y=1\}$);
for the absolute loss game, the AA 
is $(c,\eta)$-realizable for $\eta>0$
and $c\ge \eta/(2\ln(2/(1+\e^{-\eta})))$.

A detailed survey of the AA, its properties,
attainable bounds and realizability conditions 
for a number of games can be found in~\cite{Vovk:1998}.
Here we reproduce the proof of the main loss bound
in the form that motivates our further study.

\begin{theorem}[\cite{Vovk:1990}]\label{thm:AAVovk}
If the AA is $(c,\eta)$-realizable 
then the AA with parameters $c$, $\eta$, $P_0$,
and $\sigma$
guarantees that,
at each step $N$ and for all experts $\theta$, it holds
$$
 L_N\le c L_N^\theta + \frac{c}{\eta}\ln\frac{1}{P_0(\theta)}\,.
$$
\end{theorem}
\begin{proof}
We need to deduce the performance bound from
the condition~\eqref{eq:AA}.
To this end, we will rewrite~\eqref{eq:AA}
and get a semi-invariant of the AA---a value that does not grow.
Indeed, the inequality~\eqref{eq:AA} is equivalent to
$$
\sum_{\theta\in\Theta}P_{N-1}(\theta)
\ge 
\sum_{\theta\in\Theta} P_{N-1}(\theta)
\exp(-\eta\gamma_N^\theta(\omega))
\exp\left(\frac{\eta}{c}\gamma_N(\omega)\right)\,.
$$
Multiplying both sides by
$\prod_{n=1}^{N-1}\exp\left(\frac{\eta}{c}\gamma_n(\omega_n)\right)$
(which is independent of $\theta$ and hence can
be placed under the sum),
and expanding $P_{N-1}$,
we get
\begin{multline*}
\sum_{\theta\in\Theta}P_{0}(\theta)
\prod_{n=1}^{N-1}\exp(-\eta\gamma_n^\theta(\omega_n))
\prod_{n=1}^{N-1}\exp\left(\frac{\eta}{c}\gamma_n(\omega_n)\right)
\\
\ge 
\sum_{\theta\in\Theta} P_{0}(\theta)
\prod_{n=1}^{N-1}\exp(-\eta\gamma_n^\theta(\omega_n))
\prod_{n=1}^{N-1}\exp\left(\frac{\eta}{c}\gamma_n(\omega_n)\right)
\\
\times
\exp(-\eta\gamma_N^\theta(\omega))
\exp\left(\frac{\eta}{c}\gamma_N(\omega)\right)\,,
\end{multline*}
that is,
$$
\sum_{\theta\in\Theta}P_{0}(\theta) Q_{N-1}(\theta)
\ge 
\sum_{\theta\in\Theta} P_{0}(\theta)
Q_{N-1}(\theta)
\exp\left(\eta\left(
\frac{\gamma_N(\omega)}{c}-\gamma_N^\theta(\omega)
\right)\right)
$$
where $Q_{N-1}$ is defined by the formula:
$$
Q_{N-1}(\theta)=
\exp\left(\eta\sum_{n=1}^{N-1}
\biggl(\frac{\gamma_n(\omega_n)}{c}-\gamma_n^\theta(\omega_n)\biggr)\right)\,.
$$
That is, the condition~\eqref{eq:AA} 
is equivalent to
\begin{equation}\label{eq:AAtransf}
\exists \gamma_N\in\Lambda\;\forall\omega\quad
\sum_{\theta\in\Theta}P_0(\theta)\tilde Q_{N}(\theta)
\le
\sum_{\theta\in\Theta}P_0(\theta)Q_{N-1}(\theta)\,,
\end{equation}
where $\tilde Q_N$ is the result of substituting 
$\omega$ for $\omega_N$ in $Q_N$.

In other words, 
the AA (when it is $(c,\eta)$-realizable) guarantees that 
after each step $n$ the value
$\sum_{\theta\in\Theta}P_0(\theta)Q_{n}(\theta)$ does not increase
whatever $\omega_n$ is chosen by Reality.
Since 
$\sum_{\theta\in\Theta}P_0(\theta)Q_{0}(\theta)=
\sum_{\theta\in\Theta}P_{0}(\theta)=1$, we get 
$\sum_{\theta\in\Theta}P_0(\theta)Q_{N}(\theta)\le 1$ and
$Q_N(\theta)\le 1/P_0(\theta)$ for each step $N$.
To complete the proof
it remains to note that
\begin{equation*}
Q_N(\theta)=
\exp\left(\eta\left(\frac{L_N}{c}-L_N^\theta\right)\right)\,.
\end{equation*}
\end{proof}

For $c=1$,
the value $\frac{1}{\eta}\ln\left(\sum_\theta P_0(\theta)Q_N(\theta)\right)$
is known as the exponential potential 
(see~\cite[Sections~3.3,~3.5]{CBL:2006})
and plays an important role in the analysis of 
weighted average algorithms.
In the next section we show that
the reason why condition~\eqref{eq:AAtransf} 
can be satisfied is essentially that 
the function~$\sum_\theta P_0(\theta)Q_N(\theta)$
is a supermartingale.

\section{Supermartingales and the AA}
\label{sec:super}

Let $\PPP(\Omega)$ be the set of all distributions on $\Omega$.
Note that since $\Omega$ is finite we can identify
$\PPP(\Omega)$ with a $(\lvert\Omega\rvert-1)$-dimensional simplex
in Euclidean space $\mathbb{R}^{\lvert\Omega\rvert}$
equipped with the standard distance and topology.
Let $E$ be any non-empty set.
A real-valued function $S$ defined on $(E\times\PPP(\Omega)\times\Omega)^*$
is called a (game-the\-o\-re\-tic) \emph{supermartingale} 
if for any~$N$,
for any $e_1,\ldots,e_N\in E$,
for any $\pi_1,\ldots,\pi_N\in \PPP(\Omega)$,
for any $\omega_1,\ldots,\omega_{N-1}\in \Omega$,
it holds that
\begin{multline}\label{eq:super}
\sum_{\omega\in\Omega}
\pi_N(\omega) 
   S(e_1,\pi_1,\omega_1,\ldots,
     e_{N-1},\pi_{N-1},\omega_{N-1},
     e_N,\pi_N,\omega)\\
\le
S(e_1,\pi_1,\omega_1,\ldots,e_{N-1},\pi_{N-1},\omega_{N-1})\,.
\end{multline}
For $N=1$, the argument of $S$ in the right-hand side 
is the empty sequence,
and we treat $S()$ as a real constant.
The intuition behind the definition
is the following:
there is a sequence of events~$\omega_n$,
each event is generated according its own distribution~$\pi_n$
selected (or revealed) at each step anew;
when the event happens
we compute the next value of~$S$ depending on the outcomes
of the previous events,
the previous distributions and some side information~$e_n$;
the supermartingale property of $S$ means that 
the expectation of the next value (when the distribution $\pi_n$
has been selected but the outcome is not known yet)
never exceeds the previous value of~$S$.

\begin{remark}
The notion of a supermartingale is well-known in the probability theory.
Let $X_1,X_2,\ldots$ be a sequence of random elements with values
in~$\Omega$.
Denote by $x_n$ some realization of $X_n$, $n=1,2,\ldots$,
and let $\pi_n$ be a conditional distribution of $X_n$ given 
$X_1=x_1,\ldots,X_{n-1}=x_{n-1}$.
If we fix some values for $e_n$
and substitute $X_n$ for $\omega_n$ in $S$,
we can rewrite condition~\eqref{eq:super} as
$$
 \EXP S(x_1,\ldots,x_{N-1},X_N)
 \le
  S(x_1,\ldots,x_{N-1})
$$
(the parameters $e_n$ and $\pi_n$ in $S$ are omitted).
We get the usual definition of a (probabilistic) supermartingale 
$S_N=S(X_1,\ldots,X_N)$, $N=1,2,\ldots$,
with respect to the sequence $X_1,X_2,\ldots$:
$$
 \EXP [S_N\mid X_1,\ldots,X_{N-1}]
 \le
  S_{N-1}\,.
$$
In a sense, a game-theoretic supermartingale is 
a family of probabilistic supermartingales 
parameterized by some $e_n$ and also by probabilistic distributions~$\pi_n$, 
where the latter serve as conditional probabilities
of the underlying random process.
\end{remark}

\begin{remark}
A reader familiar with the supermartingales
in algorithmic probability theory 
may also find helpful the following connection. 
Let $\mu\colon\Omega^*\to[0,1]$ be a measure on $\Omega^\infty$
(where $\Omega^*$ and $\Omega^\infty$ are the sets of finite and 
infinite sequences of elements from~$\Omega$).
As defined in e.\,g.~\cite[p.~296]{LiVitanyi},
a function $s\colon\Omega^*\to\mathbb{R}_{+}$
is called a supermartingale with respect to $\mu$ if
for any $N$ and any $\omega_1,\ldots,\omega_{N-1}\in \Omega$ 
it holds that
$$
 \sum_{\omega\in\Omega}
\mu(\,\omega\mid\omega_1,\ldots,\omega_{N-1})
s(\omega_1,\ldots,\omega_{N-1},\omega)
\le
s(\omega_1,\ldots,\omega_{N-1})\,,
$$
where $\mu(\,\omega\mid\omega_1,\ldots,\omega_{N-1})
=\frac{\mu(\omega_1,\ldots,\omega_{N-1},\omega)}
{\mu(\omega_1,\ldots,\omega_{N-1})}$
(and $\mu(\omega_1,\ldots,\omega_{n})$ means the measure
of the set of all infinite sequences with the prefix 
$\omega_1\ldots\omega_{n}$).
Let $e_n$ be any functions of $\omega_1,\ldots,\omega_{n-1}$.
Let $\pi_n(\omega)$ be 
$\mu(\,\omega\mid\omega_1,\ldots,\omega_{n-1})$.
Having substituted these functions in any game-theoretic
supermartingale $S$,
we get a supermartingale with respect to $\mu$ 
in the algorithmic sense.
\end{remark}

A supermartingale $S$ is called \emph{forecast-continuous}
if for any~$N$,
for any $e_1,\ldots,e_N\in E$,
for any $\pi_1,\ldots,\pi_{N-1}\in \PPP(\Omega)$,
for any $\omega_1,\ldots,\omega_{N-1},\omega_N\in \Omega$,
the function $S(e_1,\pi_1,\omega_1,\ldots,e_N,\pi,\omega_N)$
is continuous as the function of $\pi\in\PPP(\Omega)$.

The main use of forecast-continuous supermartingales
in this paper is explained by the following lemma.

\begin{lemma}\label{lem:LevinSuper}
Suppose that $S$ is a forecast-continuous supermartingale.
Then for any $N$,
for any $e_1,\ldots,e_N\in E$,
for any $\pi_1,\ldots,\pi_{N-1}\in \PPP(\Omega)$,
for any $\omega_1,\ldots,\omega_{N-1}\in \Omega$,
it holds that
\begin{multline*}
\exists\pi\in\PPP(\Omega)\,
\forall\omega\in\Omega\quad
S(e_1,\pi_1,\omega_1,\ldots,e_N,\pi,\omega)
\le\\
S(e_1,\pi_1,\omega_1,\ldots,e_{N-1},\pi_{N-1},\omega_{N-1})\,.
\end{multline*}
\end{lemma}

Note that the property provided by this lemma
is similar to the condition~\eqref{eq:AAtransf},
where the role of $S$ with the first~$N-1$ triples of 
the arguments 
is played by 
$\sum_{\theta\in\Theta}P_0(\theta)Q_{N-1}(\theta)$,
the role of $S(\ldots,e_N,\pi,\omega)$
(the left-hand side)
is played by 
$\sum_{\theta\in\Theta}P_0(\theta)\tilde Q_{N}(\theta)$,
the variable $\pi$ corresponds to $\gamma_N$,
and for $n=1,\ldots,N-1$,
the parameters $\pi_n$ and $e_n$
are represented by $\gamma_n$ and the vector of 
$\gamma_n^\theta$, $\theta\in\Theta$, respectively. 

A variant of this lemma was
originally proved by Levin~\cite{levin:1976uniform}
in the context of algorithmic theory of randomness.
We will prove this lemma later 
(see Lemma~\ref{lem:LevinProperty}),
and in the next subsection we 
consider the Defensive Forecasting Algorithm,
the main application of this lemma in our paper.

\subsection{Defensive Forecasting}
\label{ssec:df}

The \emph{Defensive Forecasting} Algorithm (DFA)
is another strategy for Learner in 
the game of prediction with expert advice.
Let $(\Omega,\Lambda)$ be a game.
The DFA has five parameters: reals $c\ge 1$, $\eta>0$, 
a (canonic) loss function 
$\lambda\colon\PPP(\Omega)\to\Sigma_\Lambda$,
a distribution $P_0$ on $\Theta$,
and a substitution function 
$\sigma\colon\Sigma_\Lambda\to\Lambda$
such that $\sigma(\gamma)\le\gamma$ for all 
$\gamma\in\Sigma_\Lambda$.

Given $\lambda$, $c$ and $\eta$,
let us define the following function on 
$(\Sigma_\Lambda\times\PPP(\Omega)\times\Omega)^*$:
\begin{equation}\label{eq:defsuper}
Q(g_1,\pi_1,\omega_1,\ldots,g_N,\pi_N,\omega_N)=
\exp\left(\eta\sum_{n=1}^{N}
\biggl(\frac{\lambda(\pi_n,\omega_n)}{c}-
       g_n(\omega_n)\biggr)\right)\,.
\end{equation}
To simplify notation, here and in the sequel
we consider $\lambda$ as a function 
from $\PPP(\Omega)\times\Omega$ to $[0,\infty]$,
that is, we write $\lambda(\pi,\omega)$
instead of $\bigl(\lambda(\pi)\bigr)(\omega)$
and $\lambda(\pi,\cdot)$ 
instead of $\lambda(\pi)$.
For $N=0$, we let $Q()=1$ 
in accordance with the usual agreement that 
the sum of zero number of terms equals~$0$.
Note that $Q$ is similar to $Q_N(\theta)$ from the proof of 
Theorem~\ref{thm:AAVovk},
with $g_n$ standing for $\gamma^\theta_n$ and
$\lambda(\pi_n,\cdot)$ standing for $\gamma_n$.

Given also $P_0$,
let us define the function $Q^{P_0}$
on $((\Sigma_\Lambda)^\Theta\times\PPP(\Omega)\times\Omega)^*$
as the following weighted sum of $Q$:
\begin{multline}\label{eq:defcompletesuper}
Q^{P_0}(\{\gamma_1^\theta\}_{\theta\in\Theta},\pi_1,\omega_1,\ldots,
       \{\gamma_N^\theta\}_{\theta\in\Theta},\pi_N,\omega_N)
=\\
\sum_{\theta\in\Theta}
P_0(\theta)Q(\gamma_1^\theta,\pi_1,\omega_1,\ldots,
             \gamma_N^\theta,\pi_N,\omega_N)\,.
\end{multline}
At step $N$, the DFA chooses any $\pi_N\in\PPP(\Omega)$
such that 
\begin{multline}\label{eq:DFrealiz}
\forall\omega\in\Omega\quad
Q^{P_0}(\{\gamma_1^\theta\}_{\theta\in\Theta},\pi_1,\omega_1,\ldots,
        \{\gamma_N^\theta\}_{\theta\in\Theta},\pi_N,\omega)
\le\\
Q^{P_0}(\{\gamma_1^\theta\}_{\theta\in\Theta},\pi_1,\omega_1,\ldots,
        \{\gamma_{N-1}^\theta\}_{\theta\in\Theta},\pi_{N-1},\omega_{N-1})\,,
\end{multline}
stores this $\pi_N$ for use at later steps,
and announces 
$\gamma_N=\sigma(\lambda(\pi_N,\cdot))$ as Learner's prediction.

Assume that 
the function $Q$ defined by~\eqref{eq:defsuper}
is a fore\-cast-con\-ti\-nuous supermartingale.
Clearly, this implies that $Q^{P_0}$ 
defined by~\eqref{eq:defcompletesuper} 
is also a forecast-continuous supermartingale
for any $P_0$.
Then Lemma~\ref{lem:LevinSuper} guarantees that the DFA 
can choose $\pi_N$ with the required property.

\begin{theorem}\label{thm:superbound}
If $Q$ defined by~\eqref{eq:defsuper} 
is a forecast-continuous supermartingale for 
certain $c$, $\eta$, and $\lambda$
then the DFA with parameters $c$, $\eta$, $\lambda$, $P_0$,
and $\sigma$ guarantees that,
at each step $N$ and for all experts $\theta$, it holds
$$
 L_N\le c L_N^\theta + \frac{c}{\eta}\ln\frac{1}{P_0(\theta)}\,.
$$
\end{theorem}
\begin{proof}
The step of the DFA guarantees that
at each step $N$ the value of $Q^{P_0}$ does not increase 
independent of the outcome $\omega_N$. 
Thus, the value of $Q^{P_0}$ at each step~$N$
is not greater than its initial value, $1$.
Since $Q$ is always non-ne\-ga\-ti\-ve and $Q^{P_0}$
as the sum of non-negative values can be bounded from below
by any of its terms, we get
$$
P_0(\theta)\exp\left(\eta\sum_{n=1}^{N}
\biggl(\frac{\lambda(\pi_n,\omega_n)}{c}-
       \gamma^\theta_n(\omega_n)\biggr)\right)\le 1\,,
$$
and therefore
$$
\sum_{n=1}^{N}\lambda(\pi_n,\omega_n)\le 
 cL_N^\theta + \frac{c}{\eta}\ln\frac{1}{P_0(\theta)}\,.
$$
It remains to recall that 
$\gamma_n=\sigma(\lambda(\pi_n,\cdot))\le\lambda(\pi_n,\cdot)$,
thus summing up we get 
$L_N\le\sum_{n=1}^{N}\lambda(\pi_n,\omega_n)$.
\end{proof}

In Subsections~\ref{ssec:DFtoAA}--\ref{ssec:constrproper}
we discuss general conditions 
when $Q$ defined by~\eqref{eq:defsuper} is a supermartingale.
In the next subsection we begin with examples 
for two widely used games of prediction.

\subsection{Two Examples of Supermartingales}
\label{ssec:examples}

The \emph{logarithmic loss game} is defined by
the loss function
\begin{equation*}
  \lambda^\mathrm{log}(p,\omega)
  :=
  \begin{cases}
    -\ln p & \text{if $\omega=1$},\\
    -\ln(1-p) & \text{if $\omega=0$},
  \end{cases}
\end{equation*}
where $\omega\in\{0,1\}$ is the outcome and 
$p\in[0,1]$ is the decision
(notice that the loss function is allowed to take value $\infty$).
It is known~\cite[Example~5]{Vovk:1998}
that this game is $\eta$-mixable for $\eta\in(0,1]$.
The corresponding prediction set is 
$\Lambda^\mathrm{log}={\{(x,y)\in\mathbb{R}^2\mid \e^{-x}+\e^{-y}=1\}}$.
The losses in the game are
$L_N:=\sum_{n=1}^N\lambda^\mathrm{log}(p_n,\omega_n)$ 
for Learner who predicts $p_n$ and
$L_N^\theta:=\sum_{n=1}^N\lambda^\mathrm{log}(p_n^\theta,\omega_n)$
for Expert~$\theta$ who predicts $p_n^\theta$. 
Consider the following function:
\begin{equation}\label{eq:logloss}
    \exp
    \left(\eta\sum_{n=1}^N
       \Bigl(\lambda^\mathrm{log}(p_n,\omega_n)-\lambda^\mathrm{log}(p^\theta_n,\omega_n)\Bigr)
    \right)\,.
\end{equation}
This function is actually $Q$ defined by~\eqref{eq:defsuper},
where $c=1$ and $\lambda^\mathrm{log}(p^\theta_n,\cdot)$ stands for $g_n$.
The only difference is that $p_n$ is not an element of $\PPP(\Omega)$.
To fix this,
let us assign Learner's decision $p\in[0,1]$
(and thereby prediction ${(-\ln(1-p),-\ln p)}\in\Lambda^\mathrm{log}$)
to each distribution $\pi=(1-p,p)$ on $\{0,1\}$.
With this identification $\pi\mapsto p$,
the expression~\eqref{eq:logloss} specifies a function
on ${([0,1]\times\PPP(\{0,1\})\times\{0,1\})^*}$
with the arguments $p_n^\theta$, 
$\pi_n$ (represented by $p_n=\pi_n(1)$) and $\omega_n$.

\begin{lemma}\label{lem:supermartingale-log}
For $\eta\in(0,1]$,
the function~\eqref{eq:logloss} is a forecast-continuous supermartingale.
\end{lemma}
\begin{proof}
The continuity is obvious. For the supermartingale
property, it suffices to check that
  \begin{equation}\label{eq:loglossinequality}
    p_n \e^{\eta\left(-\ln p_n + \ln p_n^\theta\right)}
    +
(1-p_n) \e^{\eta\left(-\ln(1-p_n) + \ln\left(1-p_n^\theta\right)\right)}
    \le
    1
  \end{equation}
  i.e., that
$p_n^{1-\eta} \left(p_n^\theta\right)^{\eta} + 
(1-p_n)^{1-\eta} \left(1-p_n^\theta\right)^{\eta} \le 1 $
  for all $p_n,p_n^\theta,\eta\in[0,1]$.
The last inequality immediately follows from the generalized
inequality between arithmetic and geometric means: 
$u^\alpha v^{1-\alpha}\le \alpha u+ (1-\alpha) v$ for 
any $u,v\ge0$ and $\alpha\in[0,1]$,
which after taking the logarithm
just expresses that logarithm is concave.
(Remark:
The left-hand side of~\eqref{eq:loglossinequality} is a special case
of what is known as the Hellinger integral in probability theory.)
\end{proof}

In the \emph{square loss game},
the outcomes are $\omega\in\{0,1\}$ and 
the decisions are $p\in[0,1]$ as before,
and the loss function is 
$\lambda^\mathrm{sq}(p,\omega)=(p-\omega)^2$.
It is known~\cite[Example~4]{Vovk:1998}
that this game is $\eta$-mixable for $\eta\in(0,2]$.
The corresponding prediction set is 
$\Lambda^\mathrm{sq}={\{(x,y)\in[0,1]^2\mid \sqrt{x}+\sqrt{y}=1\}}$.
The losses of Learner and Expert $\theta$ are
$L_N:=\sum_{n=1}^N(p_n-\omega_n)^2$
and $L_N^\theta:=\sum_{n=1}^N(p_n^\theta-\omega_n)^2$,
respectively.
With the same identification $\pi\mapsto p$,
the following expression specifies a function
on ${([0,1]\times\PPP(\{0,1\})\times\{0,1\})^*}$:
\begin{equation}\label{eq:squareloss}
    \exp\left(
          \eta\sum_{n=1}^N
            \left((p_n - \omega_n)^2-(p^\theta_n - \omega_n)^2\right)
    \right)
\end{equation}
(again, note that it is a special case of $Q$ defined by~\eqref{eq:defsuper}).

\begin{lemma}\label{lem:supermartingale-quadratic}
For $\eta\in(0,2]$,
the  function~\eqref{eq:squareloss}
is a forecast-continuous supermartingale.
\end{lemma}
\begin{proof}
It is sufficient to check that
  $$
    p_n\e^{\eta\left((p_n - 1)^2 - (p^\theta_n - 1)^2\right)}
    +
(1-p_n)\e^{\eta\left((p_n - 0)^2 - (p^\theta_n - 0)^2\right)}
    \le
    1
  $$
for all $p_n,p^\theta_n\in[0,1]$ and $\eta\in[0,2]$.
To simplify notation,
let us substitute $p$ for $p_n$ 
and $p+x$ for $p^\theta_n$.
Then after trivial transformations we get:
  \begin{equation*}
  p\e^{2\eta(1-p)x}+(1-p)\e^{-2\eta px}\le
   \e^{\eta x^2},
    \quad
    \forall x\in[-p,1-p].
  \end{equation*}
The last inequality is a simple corollary of the following
well-known variant of 
Hoeffding's inequality~\cite[4.16]{hoeffding:1963}:
$$
\ln\EXP\e^{sX}\le s\EXP X + \frac{s^2(b-a)^2}{8}\,,
$$
which is true for any random variable~$X$ taking values in $[a,b]$
and for any $s\in\mathbb{R}$;
see~\cite[Lemma~A.1]{CBL:2006} for a proof.
Indeed, applying the inequality to the random variable
$X$ that is equal to $1$ with probability $p$
and to $0$ with probability $(1-p)$,  
we obtain
${p\exp(s(1-p))+(1-p)\exp(-sp)\le\exp(s^2/8)}$.
Substituting $s:=2\eta x$,
we have
$p\exp(2\eta(1-p)x)+(1-p)\exp(-2\eta px)\le
\exp(\eta^2 x^2 / 2)\le\exp(\eta x^2)$,
the last inequality assuming $\eta\le2$.
\end{proof}

\subsection{Supermartingales and the Realizability of the AA}
\label{ssec:DFtoAA}

Our next goal is to find when 
$Q$ defined by~\eqref{eq:defsuper} is a supermartingale,
depending on the parameters $c$, $\eta$ and $\lambda$.
Loosely speaking, we will show that
the AA is $(c,\eta)$-realizable if and only if 
there exists $\lambda$ such that $Q$ is a supermartingale.
More precisely,
the ``only if'' part holds for some class of games only.
For arbitrary games, 
the equivalence holds if 
we relax slightly the supermartingale definition
(see Theorem~\ref{thm:criterion}).

Let us begin with some notation.
For any functions $f\colon\Omega\to\mathbb{R}$ 
and $\pi\colon\Omega\to\mathbb{R}$
denote
$$
\EXP_\pi f :=
  \sum_{\omega\in\Omega} \pi(\omega)f(\omega)\,.
$$
Actually, this is the scalar product of $f$ 
and $\pi$ in $\mathbb{R}^\Omega$.
We will mostly use this for $\pi\in\PPP(\Omega)$;
in this case $\EXP_\pi f$ can be interpreted 
as the expectation of $f$ over distribution $\pi$.
For functions $g\in[0,\infty]^\Omega$
and $\pi\in\PPP(\Omega)$,
let 
$$
\EXP_\pi g :=
  \sum_{\omega\in\Omega,\: \pi(\omega)\ne0} \pi(\omega)g(\omega)\,.
$$

Recall that the function $Q$ defined by~\eqref{eq:defsuper} 
is a supermartingale if 
$$
\EXP_\pi\bigl(
         Q(g_1,\pi_1,\omega_1,\ldots,g_N,\pi,\cdot)
        -Q(g_1,\pi_1,\omega_1,\ldots,g_{N-1},\pi_{N-1},\omega_{N-1})
        \bigr)\le 0
$$
for any $g_1,\pi_1,\omega_1,\ldots,g_{N-1},\pi_{N-1},\omega_{N-1},g_N$ 
and $\pi$. 
The formula~\eqref{eq:defsuper} 
can be rewritten as
$Q=\prod_{n=1}^N q_{g_n}(\pi_n,\omega_n)$,
where the functions 
$q_{g}\colon\PPP(\Omega)\times\Omega\to[0,\infty]$
are defined by the formula 
\begin{equation}\label{eq:defsuperterm}
q_g(\pi,\omega) =
\exp\left(\eta\biggl(\frac{\lambda(\pi,\omega)}{c}-
       g(\omega)\biggr)\right)
\end{equation}
for any $g\in\Sigma_\Lambda$.
Clearly, $Q$ is a supermartingale
if and only if 
$\EXP_\pi q_g(\pi,\cdot) \le 1$
for all $\pi\in\PPP(\Omega)$ and for all $g\in\Sigma_\Lambda$.

Let us say that 
a function $q\colon\PPP(\Omega)\times\Omega\to\mathbb{R}$
has the \emph{supermartingale property} if for any $\pi\in\PPP(\Omega)$
$$
\EXP_\pi q(\pi,\cdot)\le 1\,.
$$
The function $q$ is \emph{forecast-continuous} if 
for every $\omega\in\Omega$
it is continuous as the function of $\pi$.

So, $Q$ defined by~\eqref{eq:defsuper}
is a forecast-continuous supermartingale
if and only if 
the functions $q_g$ defined by~\eqref{eq:defsuperterm} 
are forecast-continuous and 
have the supermartingale property for all $g\in\Sigma_\Lambda$.
In the sequel, we will discuss the properties of~$q_g$
instead of~$Q$.
Let us begin with a variant of Lemma~\ref{lem:LevinSuper}.

\begin{lemma}\label{lem:LevinProperty}
Let a function 
$q\colon\PPP(\Omega)\times\Omega\to\mathbb{R}$
be forecast-continuous.
If for all $\pi\in\PPP(\Omega)$ it holds that
$$
 \EXP_\pi q(\pi,\cdot) \le C\,,
$$
where $C\in\mathbb{R}$ is some constant,
then 
$$
\exists\pi\in\PPP(\Omega)\,
\forall\omega\in\Omega\quad
q(\pi,\omega) \le C\,.
$$
\end{lemma}
The proof of the lemma is given in Appendix. 
Here let us illustrate the idea behind the proof.
Consider the function 
$\phi(\pi',\pi)=\EXP_{\pi'} q(\pi,\cdot)$
and assume that it has the \emph{minimax} property:
$\min_\pi\max_{\pi'}\phi(\pi',\pi)=
\max_{\pi'}\min_{\pi}\phi(\pi',\pi)$.
Looking at the right-hand side, 
note that $\min_{\pi}\phi(\pi',\pi)\le\phi(\pi',\pi')\le C$.
Let $\pi$ minimize the left-hand side,
then we get $\max_{\pi'}\EXP_{\pi'} q(\pi,\cdot)\le C$,
that is, $\EXP_{\pi'} q(\pi,\cdot)\le C$ for any $\pi'$,
which implies the statement of the lemma if we consider
distributions $\pi'$ concentrated at each $\omega$.

Note that Lemma~\ref{lem:LevinSuper} is a simple corollary 
of Lemma~\ref{lem:LevinProperty} applied to $C=0$ and
$$
q(\pi,\omega)=
S(e_1,\pi_1,\omega_1,\ldots,e_N,\pi,\omega)
-
S(e_1,\pi_1,\omega_1,\ldots,e_{N-1},\pi_{N-1},\omega_{N-1})\,.
$$

Now let us prove that 
if $q_g$ defined by~\eqref{eq:defsuperterm}
have the supermartingale property
for all $g\in\Sigma_\Lambda$
(in other words, $Q$ is a supermartingale)
then the AA is realizable.

\begin{theorem}\label{thm:superAA}
Let $\lambda$ map $\PPP(\Omega)$ to $\Sigma_\Lambda$,
and let $c\ge 1$ and $\eta>0$ be reals
such that 
$$
q_g(\pi,\omega):=
\exp\left(\eta\biggl(\frac{\lambda(\pi,\omega)}{c}-
       g(\omega)\biggr)\right)
$$
are forecast-continuous and
have the supermartingale property for all $g\in\Sigma_\Lambda$.
Then the AA is $(c,\eta)$-realizable.
\end{theorem}
\begin{proof}
Recall that the $(c,\eta)$-realizability
is equivalent to the inequality~\eqref{eq:realiz}
for any finite $G\subseteq\Sigma_\Lambda$
and for any distribution $\rho$ on $G$.
Let us consider the following function:
$$
q(\pi,\omega)=
\sum_{g\in G}\rho(g)q_g(\pi,\omega)\,.
$$
The function $q$ is forecast-continuous and
has the supermartingale property
as a non-negative weighted sum of 
forecast-continuous functions
with the supermartingale property.
By Lemma~\ref{lem:LevinProperty} applied to this $q$
and $C=1$, there exists  
$\pi\in\PPP(\Omega)$ such that 
$q(\pi,\omega)\le 1$ for all $\omega$,
that is,
$$       
\sum_{g\in G}\rho(g)\exp\left(\eta
   \biggl(\frac{\lambda(\pi,\omega)}{c}-g(\omega)\biggr)
  \right)\le 1\,.
$$
After trivial transformations,
we get the inequality~\eqref{eq:realiz}
with $\gamma(\omega)$ replaced by $\lambda(\pi,\omega)$.
It remains to note that $\lambda(\pi,\cdot)\in\Sigma_\Lambda$.
\end{proof}

\subsection{Proper Loss Functions}
\label{ssec:proper}

The functions $q_g$ defined by~\eqref{eq:defsuperterm}
have a loss function $\lambda$ as a parameter.
In this subsection,
we consider an important property of this loss function.

A function $\lambda\colon\PPP(\Omega)\times\Omega\to[0,\infty]$
is called a \emph{proper} loss function if
for all $\pi,\pi'\in\PPP(\Omega)$
$$
 \EXP_\pi \lambda(\pi,\cdot) \le  \EXP_\pi \lambda(\pi',\cdot)\,,
$$
and $\lambda$ is \emph{strictly proper} if 
for all $\pi\ne\pi'$ the inequality is strict.

The intuition behind this definition is the following.
Assume that the outcome $\omega$ is generated according
to some distribution $\pi$.
Then the expected loss $\EXP_\pi \lambda(\pi',\cdot)$ is minimal,
if the prediction $\pi'$ equals the true distribution.
Informally speaking, proper loss functions encourage
a forecaster to announce the true subjective probabilities.
In a sense, if the loss function is proper
then the predictions have a real, not just notational,
probabilistic meaning.
The proper loss functions are well-known in the Bayesian context;
see~\cite{Dav:2007} and~\cite{GR:2007}
(note that these authors consider gains, or scores,
instead of losses, so their notation differs from ours by the sign).

We say that $\lambda$ is \emph{proper with respect to a set} 
$X\subseteq[0,\infty]^\Omega$ if 
for all $\pi\in\PPP(\Omega)$, 
it holds that $\lambda(\pi,\cdot)\in X$
and for all $g\in X$ it holds that
$$
 \EXP_\pi \lambda(\pi,\cdot) \le  \EXP_\pi g
$$
(in other words, $\lambda(\pi,\cdot)\in\arg\min_{g\in X}\EXP_\pi g$).
If the inequality holds for a fixed $\pi$ and all $g\in X$,
we will say that $\lambda$ is proper at~$\pi$.
Clearly, if $\lambda$ is proper with respect to $X$
then $\lambda$ is proper in the usual sense.
The definition has a simple geometrical interpretation.
The inequality means that the set $X$ 
lies on one side of the hyperplane
$\{x\in\mathbb{R}^\Omega \mid 
\sum_{\omega\in\Omega}\pi(\omega)x(\omega) = \EXP_\pi \lambda(\pi,\cdot)\}$,
and $X$ touches the hyperplane at $\lambda(\pi,\cdot)\in X$.
That is, $\lambda(\pi,\cdot)$ is a point where $X$ touches
the supporting hyperplane with normal $\pi.$

\begin{lemma}\label{lem:superproper}
Let $\lambda$ map $\PPP(\Omega)$ to $\Sigma_\Lambda$
and $\eta>0$ be such that the functions
$$
q_g(\pi,\omega):=
\e^{\eta(\lambda(\pi,\omega)-g(\omega))}
$$
are forecast-continuous and
have the supermartingale property for all ${g\in\Sigma_\Lambda}$
(the functions 
$q_g$ are just~\eqref{eq:defsuperterm} with $c=1$).
Then $\lambda$ is a continuous
proper loss function with respect to $\Sigma_\Lambda$.
\end{lemma}
\begin{proof}
The continuity is obvious.
Since $\e^x\ge 1 + x$ for all $x\in\mathbb{R}$, we get
$$
\EXP_\pi \e^{\eta(\lambda(\pi,\cdot)-g)} 
\ge 
\EXP_\pi\bigl(1 + \eta(\lambda(\pi,\cdot)-g)\bigr)
= 1 + \eta 
    \left(\EXP_\pi\lambda(\pi,\cdot)-\EXP_\pi g\right)\,,
$$
and from the supermartingale property we have
$\EXP_\pi\lambda(\pi,\cdot)\le \EXP_\pi g$
for all $g\in\Sigma_\Lambda$
and all $\pi\in\PPP(\Omega)$,
since $\eta>0$. 
(Remark: we get the strict inequality 
$\EXP_\pi\lambda(\pi,\cdot) < \EXP_\pi g$,
if $\lambda(\pi,\omega_0)\ne g(\omega_0)$ and 
$\pi(\omega_0)\ne0$ for some $\omega_0\in\Omega$.)
\end{proof}

From Theorem~\ref{thm:superAA}
we know that the conditions of the last lemma
imply also that the game $(\Omega,\Lambda)$ 
is $\eta$-mixable.
Let us show that the converse statement holds,
i.\,e.~the properness of $\lambda$
and mixability 
are sufficient for the supermartingale property.

\begin{lemma}\label{lem:mixablesuper}
Suppose that the game $(\Omega,\Lambda)$ is $\eta$-mixable
and $\lambda\colon\PPP(\Omega)\to\Sigma_\Lambda$ 
is a proper loss function
with respect to $\Sigma_\Lambda$.
Then the functions 
$$
q_g(\pi,\omega)=\e^{\eta(\lambda(\pi,\omega)-
       g(\omega))}
$$
have the supermartingale property
for every ${g\in\Sigma_\Lambda}$.
If $\lambda$ is continuous then 
$q_g$ are forecast-continuous.
\end{lemma}
\begin{proof}
The forecast-continuity is obvious.
Assume that the supermartingale property does not hold,
in other words, 
that $\EXP_\pi\e^{\eta(\lambda(\pi,\cdot)-g)}= 1+ \delta$
for some ${\pi\in\PPP(\Omega)}$, ${g\in\Sigma_\Lambda}$
and $\delta>0$. 
For any $\epsilon>0$
consider the point 
$$
g_\epsilon=
 -\frac{1}{\eta}\ln
    \left(
      (1-\epsilon)\e^{-\eta\lambda(\pi,\cdot)}
      +
      \epsilon\e^{-\eta g}
    \right)\,.
$$
The point $g_\epsilon$ belongs to $\Sigma_\Lambda^\eta$
by the definition of $\Sigma_\Lambda^\eta$,
and $\Sigma_\Lambda^\eta=\Sigma_\Lambda$ since the game
is $\eta$-mixable,
that is, 
$g_\epsilon\in\Sigma_\Lambda$ for any $\epsilon>0$.
When $\epsilon\to0$, we have
\begin{multline*}
g_\epsilon
    = \lambda(\pi,\cdot)
    - \frac{1}{\eta} \ln\left(
          1 + \epsilon\left(\e^{\eta(\lambda(\pi,\cdot)-g)} - 1\right)
        \right)
\\
   =\lambda(\pi,\cdot)
   -\frac{\epsilon}{\eta}
      \left(\e^{\eta(\lambda(\pi,\cdot)-g)}-1\right)+O(\epsilon^2).
\end{multline*}
Taking the expectation $\EXP_\pi$, we get
$$
\EXP_\pi g_\epsilon 
= \EXP_\pi\lambda(\pi,\cdot)
 - \frac{\epsilon}{\eta}\EXP_\pi
   \left(\e^{\eta(\lambda(\pi,\cdot)-g)}-1\right)
 +O(\epsilon^2)
= 
\EXP_\pi\lambda(\pi,\cdot)-\frac{\epsilon\delta}{\eta}+O(\epsilon^2)\,,
$$
where $\delta > 0$ by our assumption.
If $\epsilon$ is sufficiently small
then
$(\delta/\eta)\epsilon > O(\epsilon^2)$
and 
$\EXP_\pi g_\epsilon < \EXP_\pi\lambda(\pi,\cdot)$,
which is impossible since $\lambda(\pi,\cdot)$
is proper with respect to~$\Sigma_\Lambda$.
\end{proof}
An alternative, more geometrical proof of the last lemma
for binary games
the reader can find in~\cite[Lemma~3]{CV:2009}.

\subsection{The Realizability of the AA and Supermartingales}
\label{ssec:AAtoDF}

Theorem~\ref{thm:superAA} shows that if 
the functions $q_g$ defined by~\eqref{eq:defsuperterm}
are fo\-re\-cast-con\-ti\-nu\-ous and have the supermartingale property
then the AA is realizable.
We want to show the converse, 
that if the AA is realizable then
one can find $\lambda$ such that
the functions $q_g$
are forecast-continuous and have the supermartingale property.
For mixable games,
we know already that a proper loss function works
(though we do not know yet whether a proper loss function exists).
In this subsection we show that we can obtain $\lambda$ in any game
if we can construct continuous proper loss functions
for mixable games.
How to do the latter and when it is possible
is discussed in the next subsection.

To state and prove the main result of this subsection,
we need two standard assumptions (see~\cite{Vovk:1998}) 
about the game $(\Omega,\Lambda)$
and some additional notation.

\begin{assumption}
\label{assump1}
$\Lambda$ is a compact subset of $[0,\infty]^{\Omega}$
(in the extended topology).
\end{assumption}

\begin{assumption}
\label{assump2}
There exists $g_{\mathrm{fin}}\in\Lambda$ 
such that $g_{\mathrm{fin}}(\omega)<\infty$ for all 
$\omega\in\Omega$.
\end{assumption}

Note that if $\Lambda$ is compact then
$\Sigma_\Lambda$ is also compact,
as well as $\Sigma_\Lambda^\eta$.
A nice feature of compact prediction sets
is that the properties of the game
are determined by the boundary of the prediction set.

For any set $X\subseteq[0,\infty]^\Omega$,
by $\Omin X$ denote the set of minimal elements of $X$:
$g_0\in\Omin X$ if and only if for any $g\in X$ the inequality
$g_0\ge g$ implies $g_0=g$.
For a compact set $X$, for every $g\in X$ there is an element 
$g_0\in\Omin X$ such that $g_0\le g$;
that is, $X\subseteq(\Omin X+[0,\infty]^\Omega)$.
Notice that $\Omin X$ is contained in the boundary 
$\partial X$ of $X$.

Since $\Sigma_\Lambda=\Sigma_\Lambda+[0,\infty]^\Omega
=\Lambda+[0,\infty]^\Omega$,
we have $\Omin\Sigma_\Lambda=\Omin\Lambda\subseteq\Lambda$.
For compact $\Lambda$, we have
$\Sigma_\Lambda
=\Omin\Sigma_\Lambda+[0,\infty]^\Omega
=\Sigma_{\Omin\Lambda}
=\Omin\Lambda+[0,\infty]^\Omega
$.
Note also that a game is $\eta$-mixable
if and only if
$\Omin\Sigma_\Lambda^\eta\subseteq\Lambda$,
since this is equivalent to
$\Sigma_\Lambda^\eta=\Sigma_\Lambda$.
A loss function is proper with respect to $\Sigma_\Lambda^\eta$
if and only if it is proper with respect to 
$\Omin\Sigma_\Lambda^\eta$.

\begin{lemma}\label{lem:proj}
Suppose that the game $(\Omega,\Lambda)$ satisfies Assumptions~\ref{assump1} 
and~\ref{assump2}
and the AA is $(c,\eta)$-realizable for this game.
Then there is a continuous mapping 
${V\colon\Sigma_\Lambda^\eta\to\partial\Sigma_\Lambda}$
such that $V(g)\le cg$ for all $g\in\Sigma_\Lambda^\eta$.
\end{lemma}
The proof is given in Appendix.
The mapping $V$ is actually the central projection
from $\Sigma_\Lambda^\eta$ into 
the superprediction set $\Sigma_\Lambda$
(which contains $c\Sigma_\Lambda^\eta$
when the AA is $(c,\eta)$-realizable).

\begin{theorem}\label{thm:superconstr}
Let the game $(\Omega,\Lambda)$ satisfy 
Assumptions~\ref{assump1} and~\ref{assump2},
the AA be $(c,\eta)$-re\-al\-izable for this game,
and $\lambda^\eta\colon\PPP(\Omega)\to\Sigma_\Lambda^\eta$ 
be a continuous proper loss function
with respect to~$\Sigma_\Lambda^\eta$.
Then for any continuous 
${\lambda\colon\PPP(\Omega)\to\partial\Sigma_\Lambda}$
such that 
${\lambda(\pi,\cdot)\le c\lambda^\eta(\pi,\cdot)}$ 
for all $\pi\in\PPP(\Omega)$, 
the functions $q_g$ defined by~\eqref{eq:defsuperterm}
are fo\-re\-cast-con\-ti\-nu\-ous 
and have the supermartingale property
for every ${g\in\Sigma_\Lambda}$;
and there exists a continuous 
${\lambda\colon\PPP(\Omega)\to\partial\Sigma_\Lambda}$
such that 
${\lambda(\pi,\cdot)\le c\lambda^\eta(\pi,\cdot)}$ 
for all $\pi\in\PPP(\Omega)$.
\end{theorem}
\begin{proof}
The forecast-continuity is obvious.
Let us check the supermartingale property,
i.\,e., that
$$
\EXP_\pi\e^{\eta\left(\frac{\lambda(\pi,\cdot)}{c}-
       g\right)}
\le 1
$$
for all $\pi\in\PPP(\Omega)$ and all $g\in\Sigma_\Lambda$.
Since $\lambda(\pi,\cdot)\le c\lambda^\eta(\pi,\cdot)$,
it suffices that
$$
\EXP_\pi\e^{\eta(\lambda^\eta(\pi,\cdot)-g)}
\le 1\,,
$$
which follows from Lemma~\ref{lem:mixablesuper}
applied to the $\eta$-mixable game $(\Omega,\Sigma_\Lambda^\eta)$
and the proper function $\lambda^\eta$
(note that $\Sigma_\Lambda\subseteq\Sigma_\Lambda^\eta$,
hence the lemma works for all $g\in\Sigma_\Lambda$).

It remains to observe that 
$\lambda(\pi,\cdot)=V(\lambda^\eta(\pi,\cdot))$,
where $V$ is defined in Lemma~\ref{lem:proj},
has the properties we need.
\end{proof}

\subsection{Construction of a Continuous Proper Loss Function}
\label{ssec:constrproper}
In this subsection,
we fix a game $(\Omega,\Lambda)$,
fix $\eta>0$,
and consider proper loss functions with respect 
to~$\Sigma_\Lambda^\eta$.
They can be interpreted also as proper loss functions
for the $\eta$-mixable game $(\Omega,\Sigma_\Lambda^\eta)$.

\begin{lemma}\label{lem:ProperUnique}
Let $\lambda_1$ and $\lambda_2$ be
functions from $\PPP(\Omega)$ to $\Sigma_\Lambda^\eta$.
Suppose that they are proper with respect to 
$\Sigma_\Lambda^\eta$ at some point $\pi\in\PPP(\Omega)$,
that is,
$\EXP_\pi\lambda_i(\pi,\cdot)\le\EXP_\pi g$, $i=1,2$, 
for all $g\in\Sigma_\Lambda^\eta$.
Then for all $\omega\in\Omega$ we have
$$
\pi(\omega)\ne 0
\quad\Rightarrow\quad
\lambda_1(\pi,\omega)=\lambda_2(\pi,\omega)\,.
$$
\end{lemma}
The proof of the lemma is given in Appendix.

Let $\PPP^\circ(\Omega)$ be the set of all non-degenerate distributions,
i.\,e.
$$               
\PPP^\circ(\Omega) = \{\pi\in\PPP(\Omega)\mid 
                   \forall\omega\in\Omega\:\:\:\pi(\omega)>0\}\,.
$$
Lemma~\ref{lem:ProperUnique} implies that 
a proper loss function is uniquely defined on~$\PPP^\circ(\Omega)$. 
The following lemma gives a more explicit specification 
of the values of a proper loss function on~$\PPP^\circ(\Omega)$.

\begin{lemma}\label{lem:ProperNonzero}
Let the game $(\Omega,\Lambda)$ satisfy Assumptions~\ref{assump1} 
and~\ref{assump2}.
Let us define function
$H\colon\mathbb{R}^\Omega\to[-\infty,\infty)$
by the formula
\begin{equation}\label{eq:ProperMin}
 H(\pi) = \min_{g\in\Sigma^\eta_\Lambda} \EXP_\pi g\,.
\end{equation}
Let $\mathcal{H}$ be the domain where $H$ is differentiable.
Then 
$\mathcal{H}\supseteq\PPP^\circ(\Omega)$,
and the components of the gradient of $H$ 
at $\pi\in\mathcal{H}\cap\PPP(\Omega)$
constitute a continuous function 
${\lambda\colon\mathcal{H}\cap\PPP(\Omega)\to\Sigma_\Lambda^\eta}$
such that $\EXP_\pi \lambda(\pi,\cdot) = H(\pi)$.
Moreover, if ${\pi\in\PPP^\circ(\Omega)}$ 
then $\lambda(\pi,\cdot)$ is the unique point
where the minimum in~\eqref{eq:ProperMin} is attained.
\end{lemma}

\begin{remark}
The function $H(\pi)$ for $\pi\in\PPP(\Omega)$ 
is known as the generalized entropy
of the game $(\Omega,\Lambda)$;
see~\cite{GD:2004}.
For the logarithmic loss game,
$H(\pi)$ becomes the Shannon entropy of $\pi$
(cf.~\eqref{eq:loglossentropy}).
It is worth mentioning that 
one can reconstruct the superprediction set~$\Sigma_\Lambda$
from the generalized entropy of the game,
and also from the predictive complexity of the game
(see~\cite{KVV:2004} for the definitions and proofs 
in the case of binary games).
\end{remark}
The proof of the lemma is given in Appendix.  
The proof is based on the fact that the function $-H(\pi)$
is convex.
Note that $\lambda(\pi,\cdot)\in\Omin\Sigma_\Lambda^\eta$
for any ${\pi\in\PPP^\circ(\Omega)}$.
Indeed, if for some 
$\pi\in\PPP^\circ(\Omega)$
we have $\lambda(\pi,\cdot)\notin\Omin\Sigma_\Lambda^\eta$
then there exists ${g\le\lambda(\pi,\cdot)}$,
$g\in\Omin\Sigma_\Lambda^\eta$ and 
$g(\omega)<\lambda(\pi,\cdot)$ for at least one $\omega$.
Since $\pi(\omega)>0$,
we get $\EXP_\pi g<\EXP_\pi \lambda(\pi,\cdot)=H(\pi)$,
which contradicts the definition of~$H$.

Recall that if a loss function $\lambda$ is proper with respect 
to~$\Sigma_\Lambda^\eta$ then $\EXP_\pi \lambda(\pi,\cdot) = H(\pi)$.
Lemma~\ref{lem:ProperNonzero} shows that on $\PPP^\circ(\Omega)$
a proper loss function $\lambda$ exists and it is unique and continuous.
Our next task is to extend $\lambda$ continuously 
from $\PPP^\circ(\Omega)$ to $\PPP(\Omega)$.
Unfortunately, this is sometimes impossible.
Consider an example.

Let $\Omega=\{1,2,3\}$, and let the prediction set be
$$
 \Lambda=\{(-\ln p,-\ln (1-p),1)
           \mid p\in[0,1]\}\,.
$$
Actually, this is the binary logarithmic loss game
with an additional dummy outcome.
This game is $1$-mixable and 
$\Sigma_\Lambda^1=\Sigma_\Lambda$.
It is easy to check that the proper loss function 
with respect to $\Sigma_\Lambda$
is given on $\PPP^\circ(\Omega)$ by the formulas
$\lambda(\pi,i)=-\ln\frac{\pi(i)}{\pi(1)+\pi(2)}$, $i=1,2$,
and $\lambda(\pi,3)=1$.
This function can be extended continuously to all $\pi$ 
such that $\pi(1)+\pi(2)\ne0$,
so we have $\lambda(\pi,\cdot)=(\infty,0,1)$
if $\pi(1)=0$ and 
$\lambda(\pi,\cdot)=(0,\infty,1)$
if $\pi(2)=0$.
However, these continuations are 
inconsistent at the point $\pi=(0,0,1)$.
Therefore, there is no continuous function
on $\PPP(\Omega)$ which is proper with respect to 
$\Sigma_\Lambda$ for this game. 

Now let us consider three examples of games
where a continuous proper (and even strictly proper) 
loss function exists. 

The first example is the Brier game (see~\cite{Vovk:2008}),
which is a generalization of the square loss game:
$$
\lambda^\mathrm{B}(\pi,\omega)=
\sum_{o\in\Omega}(\delta_\omega(o)-\pi(o))^2
$$
where $\delta_\omega(o)=1$ if $o=\omega$
and $\delta_\omega(o)=0$ if $o\ne\omega$.
For the binary game $\Omega=\{0,1\}$, 
distribution $\pi\in\PPP(\Omega)$ is pair
$(1-p,p)$ where $p\in[0,1]$,
and hence $\lambda^\mathrm{B}(\pi,\omega)=2(p-\omega)^2$,
which is twice the loss 
$\lambda^\mathrm{sq}(p,\omega)=(p-\omega)^2$ in 
the binary square loss game as defined in 
Subsection~\ref{ssec:examples}.

The Brier game is $1$-mixable,
that is, $\Sigma_{\Lambda^\mathrm{B}}^\eta=\Sigma_{\Lambda^\mathrm{B}}$
for $\eta\le 1$.
Let us calculate $H(\pi)$ defined by~\eqref{eq:ProperMin}
for $\pi\in\PPP(\Omega)$:
\begin{multline*}
H^\mathrm{B}(\pi)
=
\min_{g\in\Sigma^\eta_{\Lambda^\mathrm{B}}} \EXP_\pi g
=\min_{g\in\Lambda^\mathrm{B}}\EXP_\pi g
=\min_{\pi'\in\PPP(\Omega)}\EXP_\pi \lambda^\mathrm{B}(\pi',\cdot)
\\
=\min_{\pi'\in\PPP(\Omega)}
  \sum_{\omega\in\Omega}\pi(\omega)
           \sum_{o\in\Omega}(\delta_\omega(o)-\pi'(o))^2
\\
=1-\sum_{\omega\in\Omega}\pi^2(\omega) +
  \min_{\pi'\in\PPP(\Omega)}
    \sum_{\omega\in\Omega}(\pi'(\omega)-\pi(\omega))^2
=1-\sum_{\omega\in\Omega}\pi^2(\omega)\,.
\end{multline*}
Clearly, $H^\mathrm{B}(\pi)$ is differentiable
on~$\PPP(\Omega)$,
hence a continuous proper loss function
for the Brier game can be computed as 
the gradient of $H^\mathrm{B}$
by Lemma~\ref{lem:ProperNonzero}.
However, it is easier to note that 
the minimum of $\EXP_\pi \lambda^\mathrm{B}(\pi',\cdot)$
is attained at $\pi'=\pi$ only,
and thus the standard form of the loss function
$\lambda^\mathrm{B}$ is proper.

\begin{remark}
Note that in the example above we computed the value of $H(\pi)$
assuming that $\pi\in\PPP(\Omega)$.
If we want to compute 
$\lambda(\pi,\omega)$
as the partial derivatives of $H(\pi)$
with respect to $\pi(\omega)$,
we must consider $H(\pi)$ 
as a function on $\mathbb{R}^\Omega$
(as stated in Lemma~\ref{lem:ProperNonzero}).
To this end, just note that $H$ is homogeneous: 
\begin{equation}\label{eq:homogen}
H(\pi)=H\left(\frac{\pi}{\sum_{\omega\in\Omega}\pi(\omega)}\right)
            \sum_{\omega\in\Omega}\pi(\omega)
\end{equation}
for $\pi\in\mathbb{R}^\Omega$.
In the Brier game example we have 
$$
H^\mathrm{B}(\pi)=
  \left(
    1-\frac{\sum_{\omega\in\Omega}\pi^2(\omega)}
           {\bigl(\sum_{\omega\in\Omega}\pi(\omega)\bigr)^2}
  \right)
  \sum_{\omega\in\Omega}\pi(\omega)\,,
$$ 
and the partial derivatives are
$$
1 - 2\pi(\omega) +\sum_{o\in\Omega}\pi^2(o)
=\lambda^\mathrm{B}(\pi,\omega)
$$
for any $\pi\in\PPP(\Omega)$.
In general, if we have a function 
$\phi\colon\mathbb{R}^\Omega\to\mathbb{R}$ such that
$\phi(\pi)=H(\pi)$ for all $\pi\in\PPP(\Omega)$,
taking the derivatives of~\eqref{eq:homogen}
we get that the proper loss function $\lambda$
can be computed by the following formula
for any~${\pi\in\PPP(\Omega)}$:
$$
\lambda(\pi,\omega)=
\phi(\pi) - \sum_{o\in\Omega}\pi(o)\phi'_{o}(\pi) + \phi'_{\omega}(\pi)\,,
$$
where $\phi'_{\omega}$ is the partial derivative of $\phi$
with respect to $\pi(\omega)$.
This formula is known from the Savage theorem~\cite{Sav:1971}
(see also~\cite[Theorem~3.2]{GR:2007};
recall that they consider scores, or gains, $-\lambda$
instead of losses $\lambda$).
\end{remark}

The second example is the Hellinger game:
$$
\lambda^\mathrm{H}(\pi,\omega)=
\frac{1}{2}\sum_{o\in\Omega}
  \left(\sqrt{\delta_\omega(o)}-\sqrt{\pi(o)}\right)^2\,.
$$
Similarly to the Brier game, we can find that
$$
H^\mathrm{H}(\pi)=
\min_{\pi'\in\PPP(\Omega)}\sum_{\omega\in\Omega}\pi(\omega)
\left(1-\sqrt{\pi'(\omega)}\right)
=\sum_{\omega\in\Omega}\pi(\omega)-\sqrt{\sum_{\omega\in\Omega}\pi^2(\omega)}\,.
$$
Here the minimum is not attained at $\pi'=\pi$
and $\lambda^\mathrm{H}$ is not proper.
Taking the derivatives, we find a proper loss function for the Hellinger game:
$$
\lambda(\pi,\omega) = 1 - \frac{\pi(\omega)}{\sqrt{\sum_{\omega\in\Omega}\pi^2(\omega)}}\,.
$$
This loss function is known as the spherical loss.

The spherical loss and the Hellinger loss specify
the same game but under different parameterization.
For binary games, this kind of ``reparameterization''
was considered in~\cite[Section~3.1]{Haussler:1998},
where a proper function $\lambda(\pi,\cdot)$ 
was called a Bayes-optimal prediction for bias~$\pi$.
More precisely, the paper~\cite{Haussler:1998}
discusses binary games specified by a loss function
$\lambda(\gamma,\omega)$, where $\omega$ is $0$ or $1$
and $\gamma\in[0,1]$.
Their Lemma~3.5 states conditions (on derivatives of $\lambda$
as a function of~$\gamma$)
when there exists 
a unique $\gamma_p$ that minimizes 
$(1-p)\lambda(\gamma,0)+p\lambda(\gamma,1)$
for each $p\in[0,1]$.
This~$\gamma_p$ can be obtained from Equation~(3.8)
in~\cite{Haussler:1998}:
\begin{equation*}
(1-p)\left.\frac{d}{d\gamma}\lambda(\gamma,0)\right|_{\gamma=\gamma_p}
+
   p \left.\frac{d}{d\gamma}\lambda(\gamma,1)\right|_{\gamma=\gamma_p}
=0\,.
\end{equation*}
Our Lemma~\ref{lem:ProperNonzero}
can be regarded as a generalization of this approach.

Our third example is the general logarithmic loss game defined by
$$
\lambda^\mathrm{log}(\pi,\omega) = -\ln\pi(\omega)\,.
$$
Similarly to the Brier loss function,
the logarithmic loss function is strictly proper.
Indeed, let us calculate 
the entropy $H^\mathrm{log}$ for $\pi\in\PPP(\Omega)$:
\begin{multline}\label{eq:loglossentropy}
H^\mathrm{log}(\pi)
=\min_{\pi'\in\PPP(\Omega)}
  \sum_{\omega\in\Omega}\pi(\omega)\bigl(-\ln\pi'(\omega)\bigr)\\
= -\sum_{\omega\in\Omega}\pi(\omega)\ln\pi(\omega)
- \max_{\pi'\in\PPP(\Omega)} 
    \sum_{\omega\in\Omega}\pi(\omega)\ln\frac{\pi'(\omega)}{\pi(\omega)}
= -\sum_{\omega\in\Omega}\pi(\omega)\ln\pi(\omega)\,.
\end{multline}
Here the partial derivatives are infinite at the bound of 
$\PPP(\Omega)$.
Nevertheless, it is easy to check that 
the minimum in the definition of $H^\mathrm{log}(\pi)$
is always attained at one point $\pi'=\pi$ only.
The last equality in~\eqref{eq:loglossentropy}
holds since logarithm is concave
$\sum_{\omega\in\Omega}\pi(\omega)\ln\frac{\pi'(\omega)}{\pi(\omega)}\le
\ln\left(\sum_{\omega\in\Omega}
        \pi(\omega)\frac{\pi'(\omega)}{\pi(\omega)}\right)
=0
$
and the inequality is strict unless 
$\frac{\pi'(\omega)}{\pi(\omega)}$ are equal for all $\omega\in\Omega$
or
$\pi(\omega_0)=1$ for some $\omega_0$.
In the former case, $\pi=\pi'$, since $\pi,\pi'\in\PPP(\Omega)$.
In the latter case, we get $\max_{\pi'\in\PPP(\Omega)}\ln\pi'(\omega_0)$,
which is attained if $\pi'(\omega_0)=1$,
and hence $\pi=\pi'$ too.

Now we consider a general way to construct 
proper loss functions,
even in the case when~$H$ 
is not differentiable on all $\PPP(\Omega)$.
Note that the only way to extend $\lambda$ continuously
is to define it at $\PPP(\Omega)\setminus\PPP^\circ(\Omega)$
as a limit from $\PPP^\circ(\Omega)$,
where $\lambda(\pi,\cdot)$ is defined as a point of minimum.
The following lemma proved in Appendix states
that a limit of such points is 
again a point of minimum.

\begin{lemma}\label{lem:ProperLimit}
Let $\pi_i\in\PPP(\Omega)$ and
$\gamma_i\in\Omin\Sigma_\Lambda^\eta$
be such that 
$\EXP_{\pi_i}\gamma_i = 
  \min_{g\in\Sigma_\Lambda^\eta} \EXP_{\pi_i}g$,
$i=1,2,\ldots$.
Assume that $\pi_i\to\pi$ and $\gamma_i\to\gamma$
as $i\to\infty$.
Then $\gamma\in\Omin\Sigma_\Lambda^\eta$
and 
$\EXP_{\pi}\gamma = 
  \min_{g\in\Sigma_\Lambda^\eta} \EXP_{\pi}g$.
\end{lemma}

In particular, the lemma implies that
a continuous proper loss function
exists in games where 
each minimum is attained in a unique point.
Let us formulate this assumption explicitly and 
prove the existence theorem.

\begin{assumption}
\label{assump3}
For every $\pi\in\PPP(\Omega)$ such that
$\pi(\omega_1)=0$ and $\pi(\omega_2)=0$
for some $\omega_1,\omega_2\in\Omega$,
$\omega_1\ne\omega_2$,
there exists only one point where 
the minimum of $\EXP_\pi g$ over all $g\in\Omin\Sigma_\Lambda^\eta$
is attained.
\end{assumption}

\begin{remark}
Assumption~\ref{assump3} holds automatically for all binary games.
The games with differentiable~$H$,
such as the general square loss game, satisfy Assumption~\ref{assump3}
as well.
\end{remark}

\begin{theorem}\label{thm:proper}
Suppose that
the game $(\Omega,\Lambda)$ satisfies Assumptions~\ref{assump1} 
and~\ref{assump2},
and Assumption~\ref{assump3} for certain $\eta>0$.
Then there exists a continuous loss function
$\lambda\colon\PPP(\Omega)\to\Omin\Sigma_\Lambda^\eta$
that is proper, and even strictly proper,
with respect to $\Sigma_\Lambda^\eta$.
\end{theorem}
\begin{proof}
Let us show first that 
the minimum of $\EXP_\pi g$ over all $g\in\Omin\Sigma_\Lambda^\eta$
is attained at one point only for all $\pi\in\PPP(\Omega)$.
For $\pi\in\PPP^\circ(\Omega)$,
it follows from Lemma~\ref{lem:ProperUnique}.
Let $\pi\in\PPP(\Omega)$ be such that
$\pi(\omega_0)=0$ for some $\omega_0\in\Omega$ and
$\pi(\omega)\ne 0$ for $\omega\ne\omega_0$.
Let $g_1,g_2\in\Omin\Sigma_\Lambda^\eta$ 
be any two points of minimum.
Again by Lemma~\ref{lem:ProperUnique},
$g_1(\omega)=g_2(\omega)$ for all $\omega\ne\omega_0$.
Therefore $g_1\le g_2$ or $g_1\ge g_2$
(since $g_1(\omega_0)$ and $g_2(\omega_0)$ are comparable, 
being two reals),
and the greater of them cannot belong to $\Omin\Sigma_\Lambda^\eta$.
Thus, $g_1=g_2$.
Assumption~\ref{assump3} works for all other $\pi\in\PPP(\Omega)$.

Let us take
$\lambda(\pi,\cdot)=\arg\min_{g\in\Omin\Sigma_\Lambda^\eta}\EXP_{\pi}g$
for all $\pi\in\PPP(\Omega)$.
Clearly, $\lambda$ is proper with respect to~$\Sigma_\Lambda^\eta$
(recall that every point in~$\Sigma_\Lambda^\eta$
is minorized by some point in~$\Omin\Sigma_\Lambda^\eta$).
Let us prove continuity.
Take any converging sequence $\pi_i\in\PPP(\Omega)$,
let $\pi$ be its limit,
and consider the corresponding $\lambda(\pi_i,\cdot)$.
Lemma~\ref{lem:ProperLimit} implies that all accumulation points of the set
$\{\lambda(\pi_i,\cdot)\}$ are points where 
$\min_{g\in\Omin\Sigma_\Lambda^\eta}\EXP_{\pi}g$
is attained, therefore $\lambda(\pi,\cdot)$ is the only accumulation point
and $\lambda(\pi_i,\cdot)$ converges to $\lambda(\pi,\cdot)$.
\end{proof}

\subsection{Defensive Forecasting Revisited}
\label{ssec:DFrevisited}

Let us review the results we obtained so far.
Theorems~\ref{thm:AAVovk}
and~\ref{thm:superbound}
gives us the same loss bound for a game $(\Omega,\Lambda)$,
if the AA is realizable and 
if $Q$ defined by~\eqref{eq:defsuper}
is a forecast-continuous supermartingale,
respectively.

The algorithms are very close in their internal structure.
We can say even more:
with the same parameters and inputs,
they give the same predictions, in some sense.
More precisely, two sets coincide:
the set of $\gamma_N\in\Lambda$ satisfying~\eqref{eq:AA}
and the set of $\gamma_N\in\Lambda$ such that $\gamma_N$ minorizes 
$\lambda(\pi_N,\cdot)$
for $\pi_N$ satisfying~\eqref{eq:DFrealiz}.

Both algorithms are applicable under almost the same conditions:
Theorem~\ref{thm:superAA} says that 
if $Q$ is a forecast-continuous supermartingale
then the AA is realizable;
Theorems~\ref{thm:superconstr}
and~\ref{thm:proper} show the converse 
for games satisfying Assumptions~\ref{assump1}--\ref{assump3}.

Whereas Assumptions~\ref{assump1} and~\ref{assump2} 
are standard and natural,
and the AA is usually considered only for the games
satisfying these assumptions,
Assumption~\ref{assump3} is new and quite cumbersome.
However, it turns out that 
with the help of a more complicated version of the DFA
we can get rid of Assumption~\ref{assump3}
and get a perfect equivalence between
the realizability of the AA and some 
supermartingale condition
(under the standard Assumptions~\ref{assump1} and~\ref{assump2} only).

To begin with, let us slightly relax the definitions concerning
supermartingales.
We say that 
a function $q\colon\PPP^\circ(\Omega)\times\Omega\to\mathbb{R}$
has the \emph{supermartingale property on $\PPP^\circ(\Omega)$} 
if for any $\pi\in\PPP^\circ(\Omega)$
$$
\EXP_\pi q(\pi,\cdot)\le 1\,.
$$
The function $q$ is \emph{forecast-continuous on $\PPP^\circ(\Omega)$} 
if for every $\omega\in\Omega$
it is continuous as the function of $\pi$
for all $\pi\in\PPP^\circ(\Omega)$.

\begin{lemma}\label{lem:LevinPropertyOpen}
Let a function 
$q\colon\PPP^\circ(\Omega)\times\Omega\to\mathbb{R}$
be non-negative
and fo\-re\-cast-con\-ti\-nu\-ous on $\PPP^\circ(\Omega)$.
Suppose that for all $\pi\in\PPP^\circ(\Omega)$ it holds that
$$
 \EXP_\pi q(\pi,\cdot) \le C\,,
$$
where $C\in[0,\infty)$ is some constant.
Then 
there exists a sequence $\{\pi^{(i)}\}_{i\in\mathbb{N}}$
such that $\pi^{(i)}\in\PPP^\circ(\Omega)$,
the sequence $\pi^{(i)}$ converges in~$\PPP(\Omega)$,
the sequences
$q(\pi^{(i)},\omega)$ converge for every $\omega\in\Omega$,
and
$$
\forall\omega\in\Omega\quad
\lim_{i\to\infty} q(\pi^{(i)},\omega) \le C\,.
$$
\end{lemma}
The proof of the lemma is given in Appendix
after the proof of Lemma~\ref{lem:LevinProperty}.

\begin{theorem}\label{thm:criterion}
Let the game~$(\Omega,\Lambda)$ 
satisfy Assumptions~\ref{assump1} and~\ref{assump2}.
The AA is $(c,\eta)$-realizable for this game
if and only if 
there exists $\lambda$ such that the functions $q_g$
defined by~\eqref{eq:defsuperterm}
are forecast-continuous on $\PPP^\circ(\Omega)$
and have the supermartingale property on $\PPP^\circ(\Omega)$
for all $g\in\Sigma_\Lambda$.
\end{theorem}
\begin{proof}
The ``only if'' part easily follows from
Lemma~\ref{lem:ProperNonzero} combined 
with (the proof of) Theorem~\ref{thm:superconstr}.

The ``if'' part is analogous to Theorem~\ref{thm:superAA}.
We need to prove inequality~\eqref{eq:realiz}
for any finite $G\subseteq\Sigma_\Lambda$
and for any distribution $\rho$ on $G$.
Let us consider the function
$$
q(\pi,\omega)=
\sum_{g\in G}\rho(g)q_g(\pi,\omega)\,,
$$
which is non-negative, 
forecast-continuous on $\PPP^\circ(\Omega)$, 
and has the supermartingale property on $\PPP^\circ(\Omega)$.
By Lemma~\ref{lem:LevinPropertyOpen} applied to this $q$
and $C=1$, there exist 
$\pi^{(i)}\in\PPP^\circ(\Omega)$ such that 
$$
\forall \omega\in\Omega\quad       
\lim_{i\to\infty}\sum_{g\in G}\rho(g)\exp\left(\eta
   \biggl(\frac{\lambda(\pi^{(i)},\omega)}{c}-g(\omega)\biggr)
  \right)\le 1\,.
$$
Let $\gamma^{(i)}=\lambda(\pi^{(i)},\cdot)\in\Sigma_\Lambda$.
Since $\Sigma_\Lambda$ is compact (by Assumption~\ref{assump1}),
the sequence $\gamma^{(i)}$ contains a convergent subsequence;
let $\gamma\in\Sigma_\Lambda$ be its limit.
Then 
$\sum_{g\in G}\rho(g)\exp\left(\eta\bigl(\gamma/c-g\bigr)\right)$
is a limit of the corresponding convergent subsequence of the sequence
$\sum_{g\in G}\rho(g)\exp\left(\eta\bigl(\gamma^{(i)}/c-g\bigr)\right)$,
and for every $\omega\in\Omega$
we get inequality~\eqref{eq:realiz}:
\begin{equation*}
\sum_{g\in G}\rho(g)\exp\left(\eta
   \biggl(\frac{\gamma(\omega)}{c}-g(\omega)\biggr)
  \right)\le 1\,.
\end{equation*}
\end{proof}

Let us state also the algorithm DFA$^\ast$, 
a variant of the DFA
suitable for supermartingales on $\PPP^\circ(\Omega)$.
At step~$N$, the DFA$^\ast$ defines
the function
\begin{multline*}
q(\pi,\omega)
=
\sum_{\theta\in\Theta}
P_0(\theta)
\exp\left(\eta\sum_{n=1}^{N-1}
\biggl(\frac{\gamma_n(\omega_n)}{c}-
       \gamma_n^\theta(\omega_n)\biggr)\right)
\\
\times
\exp\left(\eta
\biggl(\frac{\lambda(\pi,\omega)}{c}-
       \gamma_N^\theta(\omega)\biggr)\right)
\end{multline*}
and chooses any sequence of $\pi^{(i)}\in\PPP^\circ(\Omega)$
such that
$$
\forall\omega\in\Omega\quad
\lim_{i\to\infty} q(\pi^{(i)},\omega) \le 1\,.
$$
Then the algorithm chooses as $\gamma$ 
the limit of any convergent subsequence
of the sequence $\lambda(\pi^{(i)},\cdot)$,
and announces $\gamma_N=\sigma(\gamma)$ as Learner's prediction.
It is clear that the DFA$^\ast$
guarantees the same loss bound as Theorem~\ref{thm:superbound}.

It is important for applications that
the AA is rather efficient computationally
(though it is more complicated than some other algorithms).
The DFA$^\ast$ is designed to obtain a nice theory, and 
it makes little sense to discuss its efficiency.
The DFA is much more practical then the DFA$^\ast$.
Unfortunately, the DFA seems to be less practical than the AA.
Its main step hidden in the proof of Lemma~\ref{lem:LevinProperty}
requires finding a fixed point (or a minimax),
which is generally a hard task (PPAD-complete).
For binary games, however,
the fixed points can be found by bisection method,
which gives us a not so inefficient implementation of the DFA.
Some tricks can also help for games with three outcomes.

\begin{remark}
After this paper had been finished,
the authors have discovered another way to deal with games
that do not satisfy Assumption~\ref{assump3}. 
The idea is to consider a multivalued loss function:
to every $\pi$ it assigns all points where the minimum of 
$\EXP_\pi g$ is attained.
The definition of supermartingale 
should be modified accordingly,
and a variant of Lemma~\ref{lem:LevinSuper} 
can be proved for such multivalued supermartingales.
The details will be added later or published elsewhere.
\end{remark}

\subsection{On Continuous Outcomes}
\label{ssec:cont}

We assumed so far that the space of outcomes, $\Omega$, 
is finite.
However,
it is often natural to consider a continuous space of outcomes.
For example, 
for the square loss function 
$\lambda^\mathrm{sq}(p,\omega)=(p-\omega)^2$,
one can take $\omega\in[0,1]$ instead of $\omega\in\{0,1\}$.

In this subsection we consider
one important case of continuous outcome spaces:
a finite-dimensional simplex.
We will consider a simplex as the space $\PPP(\Omega)$ 
of distributions on some finite $\Omega$.
A game of prediction is a pair $(\PPP(\Omega),\Lambda)$,
where $\Lambda\subseteq[0,\infty]^{\PPP(\Omega)}$;
predictions are functions $\gamma\colon\PPP(\Omega)\to[0,\infty]$;
the protocol is the same.
Each game of prediction with the outcomes from a simplex 
$\PPP(\Omega)$ can be restricted to a game on $\Omega$:
we identify each $\omega\in\Omega$ with the distribution~$\delta_\omega$
concentrated on this $\omega$.
Thus we may assume $\PPP(\Omega)\supset\Omega$.
Denote by $\Lambda_\Omega\subseteq[0,\infty]^\Omega$ 
the set of functions from $\Lambda$
restricted to $\Omega$.

We will show how the supermartingale technique 
works for games having some regularity property.
(A similar extension for the AA 
is discussed in~\cite[Section~4.1]{Haussler:1998}.)

To motivate this kind of property,
let us start from the other side
and assume that we have a prediction 
(recall that our prediction 
is a vector of our losses for every possible outcome)
$\gamma$ defined on $\Omega$ and want to extend it
to $\PPP(\Omega)$.
The most natural way to do this is to say that 
an element of $\PPP(\Omega)$ is just a probability distribution
on the outcomes,
and consider the expected loss with respect to this distribution,
that is, $\gamma(p):=\EXP_p\gamma$ for every $p\in\PPP(\Omega)$.
It is also natural to expect that having this property
one should be able to transfer a regret bound 
from the game on $\Omega$ to the respective game on $\PPP(\Omega)$.
However, the equality $\gamma(p)=\EXP_p\gamma$ is too restrictive.
For example, it does not hold for the square loss.
At the same time, what does hold for the square loss
(and will be checked later) 
is an equality concerning the difference of two predictions:
$\gamma_1(p)-\gamma_2(p)=\EXP_p(\gamma_1-\gamma_2)$.
This is quite natural in our context,
since the difference is a regret,
loosely speaking,
and a regret is the value we are optimizing.
This leads to the following requirement
(formally weaker than the condition for the square loss).

We say that
$\Lambda\subseteq[0,\infty]^{\PPP(\Omega)}$
has the \emph{relative exp-convexity property} 
for certain $c$ and $\eta$
if for all $\gamma_1,\gamma_2\in\Lambda$ 
and for all $p\in\PPP(\Omega)$ it holds that
$$
\exp\left(\eta\biggl(\frac{\gamma_1(p)}{c}-\gamma_2(p)\biggr)\right)\le 
 \sum_{\omega\in\Omega} p(\omega)
 \exp\left(\eta
  \biggl(\frac{\gamma_1(\omega)}{c}-\gamma_2(\omega)\biggr)\right)\,.
$$
\begin{remark}\label{rem:expconv}
The relative exp-convexity property for any $c>0$ and $\eta$
follows from 
$$
\forall\,\gamma\in\Lambda\:
\forall\,p\in\PPP(\Omega)\quad
\gamma(p)=\sum_{\omega\in\Omega}p(\omega)\gamma(\omega)
$$
due to convexity of the exponent function.
For $c=1$ and any $\eta$,
it follows also from 
$$
\forall\,\gamma_1,\gamma_2\in\Lambda\:
\forall\,p\in\PPP(\Omega)\quad
\gamma_1(p)-\gamma_2(p)= 
 \sum_{\omega\in\Omega} p(\omega)
    \bigl(\gamma_1(\omega)-\gamma_2(\omega)\bigr)\,.
$$
\end{remark}

Let 
$\sigma_\Omega\colon\Lambda_\Omega\to\Lambda$
be any mapping inverse to the restriction from $\Lambda$
to $\Lambda_\Omega$,
that is,
for any $\gamma\in\Lambda_\Omega$,
the function $\sigma_\Omega(\gamma)\in\Lambda$
restricted to $\Omega$
is~$\gamma$.
Such a mapping exists since every element of $\Lambda_\Omega$
is a restriction of some element of $\Lambda$.

\begin{theorem}\label{thm:contsuperbound}
For a game $(\PPP(\Omega),\Lambda)$,
suppose that 
$\Lambda$ has the relative exp-con\-vex\-ity property 
for some $c\ge1$ and $\eta>0$.
For the restricted game $(\Omega,\Lambda_\Omega)$,
suppose that 
for some 
$\lambda\colon\PPP(\Omega)\to\Sigma_{\Lambda_\Omega}$,
the functions $q_g$ defined by~\eqref{eq:defsuperterm}
are fo\-re\-cast-con\-ti\-nu\-ous
and have the supermartingale property
for all ${g\in\Sigma_{\Lambda_\Omega}}$.
Let $\sigma\colon\Sigma_{\Lambda_\Omega}\to\Lambda_\Omega$
be a substitution function 
(that is, 
$\sigma(g)\le g$ for all $g\in\Sigma_{\Lambda_\Omega}$).
Then for the game $(\PPP(\Omega),\Lambda)$
there is Learner's strategy (in fact, a variant of the DFA)
with parameters $c$, $\eta$, $\lambda$, $P_0$,
$\sigma$, and $\sigma_\Omega$
guaranteeing that,
at each step $N$ and for all experts $\theta$,
it holds
$$
 L_N\le c L_N^\theta + \frac{c}{\eta}\ln\frac{1}{P_0(\theta)}\,.
$$
\end{theorem}
\begin{proof}
Assume that we are at step~$N$ 
and need to announce the next prediction.
Let $\gamma_n^\theta\in\Lambda$, $n=1,\ldots,N$
be the experts' prediction up to step~$N$,
$\gamma_n$, $n=1,\ldots,N-1$ be the Learner's previous predictions,
and $p_n$, $n=1,\ldots,N-1$ be the previous outcomes.
Define the function $Q_{N-1}$ from $\Theta$ to $\mathbb{R}$
$$
Q_{N-1}(\theta)
=
\prod_{n=1}^{N-1}
\exp\left(\eta\biggl(\frac{\gamma_n(p_n)}{c}-
       \gamma_n^\theta(p_n)\biggr)\right)
$$
and 
consider the following function on $\PPP(\Omega)\times\Omega$:
$$
q_N(\pi,\omega)
=\sum_{\theta\in\Theta}P_0(\theta)
Q_{N-1}(\theta)\times
\exp\left(\eta\biggl(\frac{\lambda(\pi,\omega)}{c}-
       \gamma_N^\theta(\omega)\biggr)\right)\,.
$$
Due to the assumptions about the last multiplier,
$q_N$ is forecast-continuous and 
$\EXP_\pi q_N(\pi,\cdot)\le 
\sum_{\theta\in\Theta}P_0(\theta)Q_{N-1}(\theta)$.
By Lemma~\ref{lem:LevinProperty},
we can find $\pi_N\in\PPP(\Omega)$ such that for all $\omega\in\Omega$
$$
q_N(\pi_N,\omega)\le \sum_{\theta\in\Theta}P_0(\theta)Q_{N-1}(\theta)\,.
$$
The prediction of the strategy is 
$\gamma_N=\sigma_\Omega(\sigma(\lambda(\pi_N,\cdot)))\in\Lambda$. 

Let $p_N\in\PPP(\Omega)$ be the outcome at step~$N$.
The relative exp-convexity property implies that
$$
\exp\left(\eta\biggl(\frac{\gamma_N(p_N)}{c}-
       \gamma_N^\theta(p_N)\biggr)\right)
\le
\sum_{\omega\in\Omega}p_N(\omega)
 \exp\left(\eta\biggl(\frac{\gamma_N(\omega)}{c}-
       \gamma_N^\theta(\omega)\biggr)\right)\,.
$$ 
We have $\gamma_N(\omega)=\sigma(\lambda(\pi_N,\cdot))(\omega)$
by the definition of $\sigma_\Omega$,
hence we have $\gamma_N(\omega)\le\lambda(\pi_N,\omega)$
by definition of $\sigma$.
Thus,
\begin{multline*}
\sum_{\theta\in\Theta}P_0(\theta)Q_{N}(\theta)
=
\sum_{\theta\in\Theta}P_0(\theta)Q_{N-1}(\theta)\times
\exp\left(\eta\biggl(\frac{\gamma_N(p_N)}{c}-
       \gamma_N^\theta(p_N)\biggr)\right)
\\
\le 
\sum_{\theta\in\Theta}P_0(\theta)Q_{N-1}(\theta)\times
\sum_{\omega\in\Omega}p_N(\omega)
 \exp\left(\eta\biggl(\frac{\lambda(\pi_N,\omega)}{c}-
       \gamma_N^\theta(\omega)\biggr)\right)
\\
=\sum_{\omega\in\Omega}p_N(\omega)q_N(\pi_N,\omega)
\le
\sum_{\theta\in\Theta}P_0(\theta)Q_{N-1}(\theta)\,,
\end{multline*}
and the loss bound follows as usual.
\end{proof}

As an example, 
let us again consider the Brier game (the general square loss function),
now with distributions as outcomes:
$\Omega$ is a finite non-empty set,
outcomes $p$ are from $\PPP(\Omega)$,
and the loss of decision $\pi\in\PPP(\Omega)$
for outcome $p$ is
$$
\lambda^\mathrm{B}(\pi,p)=\sum_{\omega\in\Omega}(p(\omega)-\pi(\omega))^2\,.
$$
It is easy to check that this game has the
relative exp-convexity property
for $c=1$ and any $\eta$
due to Remark~\ref{rem:expconv}:
\begin{multline*}
\sum_{\omega\in\Omega}p(\omega)
   \bigl(\lambda^\mathrm{B}(\pi_1,\omega)-\lambda^\mathrm{B}(\pi_2,\omega)\bigr)\\
=\sum_{\omega\in\Omega}(\pi_1^2(\omega)-\pi_2^2(\omega))+
   2\sum_{\omega\in\Omega}p(\omega)(\pi_2(\omega)-\pi_1(\omega))\\
=\lambda^\mathrm{B}(\pi_1,p)-\lambda^\mathrm{B}(\pi_2,p)\,.
\end{multline*}

Another important example is the Kullback-Leibler game
(its restricted version is the logarithmic loss game):
$$
\lambda^\mathrm{KL}(\pi,p)=\sum_{\omega\in\Omega}
   p(\omega)\ln\frac{p(\omega)}{\pi(\omega)}\,.
$$
This game also has the relative exp-convexity property
for $c=1$ and any $\eta$:
$\lambda^\mathrm{KL}(\pi_1,p)-\lambda^\mathrm{KL}(\pi_2,p)=
\sum_{\omega\in\Omega}p(\omega)
   \bigl(\lambda^\mathrm{KL}(\pi_1,\omega)-\lambda^\mathrm{KL}(\pi_2,\omega)\bigr)$.

\section{Second-Guessing Experts}
\label{sec:second-guess}

In this section, 
we apply the supermartingale technique and the DFA
to a new variant of the prediction with expert advice setting.
Protocol~\ref{prot:PEA-SG}
is an extension of Protocol~\ref{prot:PEA}, where 
the game is specified by the same elements $(\Omega,\Lambda)$ as before,
but the Experts have a new power.

\begin{algorithm}[ht]
  \caption{Prediction with Second-Guessing Expert Advice}
  \label{prot:PEA-SG}
  \begin{algorithmic}
    \STATE $L_0:=0$.
    \STATE $L_0^\theta:=0$, for all $\theta\in\Theta$.
    \FOR{$n=1,2,\dots$}
      \STATE All Experts $\theta\in\Theta$ announce 
            $\Gamma_n^\theta\colon\Lambda\to\Lambda$.
      \STATE Learner announces $\gamma_n\in\Lambda$.
      \STATE Reality announces $\omega_n\in\Omega$.
      \STATE $L_n:=L_{n-1}+\gamma_n(\omega_n)$.
      \STATE   $L_n^\theta:=L_{n-1}^\theta+\Gamma_n^\theta(\gamma_n,\omega_n)$, 
                                 for all $\theta\in\Theta$.
    \ENDFOR
  \end{algorithmic}
\end{algorithm}

The new protocol contains only one substantial change.
Every Expert~$\theta$ announces a function $\Gamma^\theta$
from $\Lambda$ to $\Lambda$
instead of an element of $\Lambda$
(to simplify notation, we consider $\Gamma$ also as a function from 
$\Lambda\times\Omega$ to $[0,\infty]$,
as we did with the proper loss functions $\lambda$).
Informally speaking,
now an expert's opinion is not a prediction,
but a conditional statement that specifies
the actual prediction depending on Learner's next step. 
Therefore, the loss of each expert is determined by
the Learner's prediction as well as by the outcome chosen by Reality.
We will call the experts in Protocol~\ref{prot:PEA-SG} 
\emph{second-guessing experts}.
Second-guessing experts are a generalization 
of experts in the standard Protocol~\ref{prot:PEA}:
a standard expert can be interpreted in Protocol~\ref{prot:PEA-SG}
as a constant function.

The phenomenon of ``second-guessing experts'' occurs,
for example, in real-world finance.
In particular, commercial banks serve as 
``second-guessing experts'' for the central bank when they 
use variable interest rates
(that is, the interest rate for the next period 
is announced not as a fixed value
but as an explicit function of the central bank base rate).

In game theory, 
the notion of internal regret~\cite{FV:1999,BM:2007,StL:2005,StL:2007}
is somewhat related to the idea of second-guessing experts.
The internal regret appears in the framework where for each prediction,
which is called action in that context,
there is an expert that consistently recommends this action,
and Learner follows one of the experts at each step.
The internal regret for a pair of experts $(i,j)$
shows by how much Learner could have decreased his loss 
if he had followed expert~$j$ each time he followed expert~$i$.
This can be modeled by a second-guessing expert that
``adjusts'' Learner's predictions:
agrees with Learner if Learner does not follow~$i$,
and recommends following~$j$ when the Learner follows~$i$.

The internal regret is usually studied 
in randomized prediction protocols.
In the case of deterministic Learner's predictions,
one cannot hope to get any interesting loss bound 
without additional assumptions. 
Indeed,  
Experts can always suggest exactly the ``opposite''
to the Learner's prediction
(for example, in the log loss game, they predict $1$ if
Learner predicts $p_n$ (``the probability of $1$'')
less than $0.5$ and they predict $0$ otherwise), 
and Reality can ``agree'' with them
(choosing the outcome equal to Experts' prediction);
then the Experts' losses remain zero, but the Learner's loss 
grows linearly in the number of steps.
A non-trivial bound is possible if Learner is allowed to give
predictions in the form of a distribution on Experts.
This can be formalized as 
the Freund-Schapire game~\cite[Example~7]{Vovk:1998}.
Then the second-guessing expert modeling an internal regret
is a continuous transformation of the distribution
given by Learner.
The results of~\cite{BM:2007} and others
are bounds of the form 
$L_N\le L_N^\theta + O(\sqrt{N})$
for the Freund-Schapire game, which is non-mixable.
A discussion of bounds of this form achievable by 
the defensive forecasting method will be published elsewhere:
in this paper we consider another kind of bounds.
However, here we will also make the assumption
that second-guessing experts 
modify the prediction of Learner \emph{continuously}.

\subsection{The DFA for Second-Guessing Experts}

First consider the case when 
$\Gamma_n^\theta$ are continuous mappings 
from $\Lambda$ to $\Lambda$.
The DFA requires virtually no modifications
for this task
and gives the same loss bounds as in Theorem~\ref{thm:superbound}.

\begin{theorem}\label{thm:supersecond}
Suppose that for some $c$, $\eta$, and some continuous
$\lambda\colon\PPP(\Omega)\to\Lambda$
the functions $q_g$ defined by~\eqref{eq:defsuperterm} 
are forecast-continuous 
and have the supermartingale property
for all $g\in\Lambda$.
Then for the game following the protocol of prediction 
with second-guessing expert advice
where all experts $\theta$ at all steps $n$ 
announce continuous functions 
$\Gamma^\theta_n\colon\Lambda\to\Lambda$,
there is Learner's strategy 
(in fact, the DFA applied to $Q^{P_0}$
defined by~\eqref{eq:defsecondsuper}) 
with parameters $c$, $\eta$, $\lambda$, $P_0$,
(where $P_0$ is a distribution on $\Theta$)
guaranteeing that,
at each step $N$ and for all experts $\theta$, it holds
$$
 L_N\le c L_N^\theta + \frac{c}{\eta}\ln\frac{1}{P_0(\theta)}\,.
$$
\end{theorem}
\begin{proof}
For any continuous $\Gamma\colon\Lambda\to\Lambda$
consider the function
$$
\tilde q_\Gamma(\pi,\omega) =
\exp\left(\eta\biggl(\frac{\lambda(\pi,\omega)}{c}-
       \Gamma(\lambda(\pi,\cdot),\omega)\biggr)\right)\,.
$$
It is forecast-continuous as a composition of continuous functions,
and has the supermartingale property since 
for any $\pi\in\PPP(\Omega)$,
taking $g=\Gamma(\lambda(\pi,\cdot))$
we have $\EXP_\pi \tilde q_\Gamma=\EXP_\pi q_g\le 1$.
Similarly to~\eqref{eq:defcompletesuper},
define $Q^{P_0}$
on ${((C(\Lambda\to\Lambda))^\Theta\times\PPP(\Omega)\times\Omega)^*}$,
where ${C(\Lambda\to\Lambda)}$ is the set of continuous 
functions on $\Lambda$,
by the formula
\begin{multline}\label{eq:defsecondsuper}
Q^{P_0}(\{\Gamma_1^\theta\}_{\theta\in\Theta},\pi_1,\omega_1,\ldots,
       \{\Gamma_N^\theta\}_{\theta\in\Theta},\pi_N,\omega_N)
=\\
\sum_{\theta\in\Theta}
P_0(\theta)\prod_{n=1}^N
\exp\left(\eta\biggl(\frac{\lambda(\pi_n,\omega_n)}{c}-
       \Gamma_n^\theta(\lambda(\pi_n,\cdot),\omega_n)\biggr)\right)\,.
\end{multline}
As in Theorem~\ref{thm:superbound},
$Q^{P_0}$ is a forecast-continuous supermartingale.

At step~$N$, the strategy chooses any $\pi_N$ 
satisfying~\eqref{eq:DFrealiz}
and announces $\gamma_N=\lambda(\pi_N,\cdot)$
as Learner's prediction
(we do not need a substitution function here since 
the range of $\lambda$ is in $\Lambda$ by the theorem assumption).
The loss bound follows, since
\begin{multline*}
\exp\left(\eta\sum_{n=1}^{N}
\biggl(\frac{\lambda(\pi_n,\omega_n)}{c}-
       \Gamma^\theta_n(\gamma_n,\omega_n)\biggr)\right)
=\\
\exp\left(\eta\sum_{n=1}^{N}
\biggl(\frac{\lambda(\pi_n,\omega_n)}{c}-
       \Gamma^\theta_n(\lambda(\pi_n,\cdot),\omega_n)\biggr)\right)
\le\frac{1}{P_0(\theta)}\,.
\end{multline*}
\end{proof}
                
Recall that Theorem~\ref{thm:proper}
provides us (under Assumptions~\ref{assump1}--\ref{assump3})
with a continuous proper loss function 
${\lambda\colon\PPP(\Omega)\to\Omin\Sigma_\Lambda^\eta}$.
For any $\eta$-mixable game, we have
$\Omin\Sigma_\Lambda^\eta\subseteq\Lambda$,
and due to Theorem~\ref{thm:superconstr}
we can take this $\lambda$ and get
forecast-continuous $q_g$ with the supermartingale property.

For non-mixable games there is no guarantee
that such $\lambda$ exists.
Theorem~\ref{thm:superconstr}
gives a function $\lambda$ ranging over~$\partial\Sigma_\Lambda$
(the boundary of the superpredictions set~$\Sigma_\Lambda$),
which is not necessarily contained in~$\Lambda$.
Moreover, it may happen that even for continuous 
experts $\Gamma_n^\theta\colon\Lambda\to\Lambda$
it is impossible to get any interesting loss bound,
for any strategy.
Indeed, consider a game where $\Lambda$ is not connected
(e.\,g., the simple prediction game~\cite[Example~1]{Vovk:1998}
with $\Lambda=\{(0,1),(1,0)\}$).
Then the example with ``opposite'' predictions works:
the experts just need to map Learner's predictions
into another connected component.

By this reason,
let us consider a modification of Protocol~\ref{prot:PEA-SG}
that changes the sets of predictions allowed for Learner 
and for Experts.
Namely, for the game $(\Omega,\Lambda)$,
Experts $\theta\in\Theta$ announce 
$\Gamma_n^\theta\colon\partial\Sigma_\Lambda\to\Sigma_\Lambda$,
and Learner announces $\gamma_n\in\partial\Sigma_\Lambda$
(the rest of Protocol~\ref{prot:PEA-SG} does not change).
We will assume that the game satisfies Assumptions~\ref{assump1} 
and~\ref{assump2}
(for non-compact $\Lambda$ the boundary $\partial\Sigma_\Lambda$
may be empty).
Then the modified protocol usually gives more freedom to Learner:
since $\Omin\Lambda\subseteq\partial\Sigma_\Lambda$,
the predictions in $\Lambda\setminus\partial\Sigma_\Lambda$ 
are minorized by some better predictions in $\Omin\Lambda$.
The Experts are allowed to give predictions 
(which are $\Gamma_n^\theta(\gamma_n)$)
in a larger set $\Sigma_\Lambda$,
however, they need to cope with Learner predictions from a larger set too.

For the modified protocol,
Theorem~\ref{thm:supersecond} holds with minimal changes:
$\lambda$ is allowed to range over $\Sigma_\Lambda$ instead of $\Lambda$,
the functions $q_g$ have the supermartingale property
for all $g\in\Sigma_\Lambda$ (instead of $g\in\Lambda$ only),
$\Gamma^\theta_n$ are continuous functions 
from $\partial\Sigma_\Lambda$ to $\Sigma_\Lambda$;
the proof does not change.
Theorem~\ref{thm:superconstr}
provides us with $\lambda$ such that $q_g$ 
have the required properties.

\subsection{The AA for Second-Guessing Experts}

In contrast to the DFA,
the AA cannot be applied to the second-guessing protocol 
in a straightforward way.
However, the AA can be modified for this case.
Recall that the AA is based on the inequality~\eqref{eq:AA},
which is already solved for~$\gamma_N$.
In the second-guessing protocol, 
both sides of this inequality will contain~$\gamma_N$:
\begin{equation*}
\gamma_N(\omega_N)\le 
 -\frac{c}{\eta}\ln\left(\sum_{\theta\in\Theta}
\frac{P_{N-1}(\theta)}{\sum_{\theta\in\Theta}P_{N-1}(\theta)}
\exp(-\eta\Gamma_N^\theta(\gamma_N,\omega_N))\right)\,.
\end{equation*}
The DFA implicitly solves this inequality in (the proof of) 
Lemma~\ref{lem:LevinSuper}, using a kind of fixed point theorem.
We will present a modification of the AA which uses 
a fixed point theorem explicitly.

A topological space $X$ has the \emph{fixed point property}
if every continuous function $f\colon X\to X$ has a fixed point,
that is, $\exists x\in X\:f(x)=x$.

Let us show that if the game $(\Lambda,\Omega)$ satisfies
Assumptions~\ref{assump1} and~\ref{assump2}
then the set $\Omin\Sigma_\Lambda^\eta$ 
(the set of minimal points of~$\Sigma_\Lambda^\eta$)
has the fixed point property for any $\eta>0$.
First consider the homeomorphism from 
$[0,\infty]^\Omega$ to $[0,1]^\Omega$ that maps
$g\mapsto \exp(-\eta g)$.
As mentioned in Section~\ref{sec:AA},
the set $\exp(-\eta\Sigma_\Lambda^\eta)$
is convex.
It is non-empty due to Assumption~\ref{assump2} and
compact due to Assumption~\ref{assump1}.
Thus,
$\exp(-\eta\Sigma_\Lambda^\eta)$
has the fixed point property by~\cite[Theorem~4.10]{agar:2001},
and $\Sigma_\Lambda^\eta$ has the property
as its homeomorphic image~\cite[Theorem~4.1]{agar:2001}.
Now we need the following technical lemma 
proved in Appendix.
\begin{lemma}\label{lem:AAcont}
There is a continuous mapping 
$F\colon\Sigma_\Lambda^\eta\to\Omin\Sigma_\Lambda^\eta$
such that $F(g)\le g$ for any $g\in\Sigma_\Lambda^\eta$.
\end{lemma}
\begin{remark}
Essentially, 
the main contents of Lemma~\ref{lem:AAcont} 
is a construction of a continuous substitution function.
In many natural games,
the standard substitution functions are continuous
without additional efforts.
\end{remark}

The definition of $\Omin\Sigma_\Lambda^\eta$
implies that if $F(g)\le g$ then $F(g)=g$
for any $g\in\Omin\Sigma_\Lambda^\eta$,
and hence $F$ defined in the lemma is a \emph{retraction}
(by definition, a continuous mapping 
from a topological space into its subset
that does not move elements of the subset).
Due to~\cite[Theorem~4.2]{agar:2001},
since $\Sigma_\Lambda^\eta$ has the fixed point property,
its retract $\Omin\Sigma_\Lambda^\eta$
has the fixed point property too.

\begin{theorem}\label{thm:AAfixed}
Suppose that the game $(\Omega,\Lambda)$
satisfies Assumptions~\ref{assump1}
and~\ref{assump2}
and is $\eta$-mixable.
Then 
for the prediction with second-guessing expert advice protocol,
there exists Learner's strategy 
(a modification of the AA)
with parameters $\eta$ and $P_0$
guaranteeing that,
at each step $N$ and for all experts $\theta$, it holds
$$
 L_N\le L_N^\theta + \frac{1}{\eta}\ln\frac{1}{P_0(\theta)}\,.
$$
\end{theorem}
\begin{proof}
At step~$N$, the modified AA 
announces as Learner's prediction~$\gamma_N$ any solution of the 
following equation with respect to $\gamma\in\Omin\Sigma_\Lambda^\eta$:
\begin{equation}\label{eq:AAfixedmix}
\gamma = F\left(-\frac{1}{\eta}\ln\left(\sum_{\theta\in\Theta}
\frac{P_{N-1}(\theta)}{\sum_{\theta\in\Theta}P_{N-1}(\theta)}
\exp(-\eta\Gamma_N^\theta(\gamma))\right)\right)\,,
\end{equation}
where $\Gamma_N^\theta$ are announced by the experts,
the weights $P_{N-1}$
are defined in the usual way with the help of the previous losses:
$$
P_{N-1}(\theta)=P_0(\theta)
\prod_{n=1}^{N-1}\exp(-\eta\Gamma_n^\theta(\gamma_n,\omega_n))\,,
$$
and $F$ is the continuous mapping from Lemma~\ref{lem:AAcont}.

Since for an $\eta$-mixable game we have 
$\Sigma_\Lambda^\eta=\Sigma_\Lambda$,
and since $\Omin\Sigma_\Lambda\subseteq\Lambda$,
the functions $\Gamma_N^\theta$
are defined on $\gamma$.
By the definition of $\Sigma_\Lambda^\eta$,
the argument of $F$ in equation~\eqref{eq:AAfixedmix}
belongs to $\Sigma_\Lambda^\eta$,
and $F$ maps it to $\Omin\Sigma_\Lambda^\eta$.
The mapping is continuous as the composition of continuous mappings.
Therefore, 
since 
$\Omin\Sigma_\Lambda^\eta$ has the fixed point property,
equation~\eqref{eq:AAfixedmix} has a solution.

The property $F(g)\le g$ implies that 
$$
\gamma_N \le -\frac{1}{\eta}\ln\left(\sum_{\theta\in\Theta}
\frac{P_{N-1}(\theta)}{\sum_{\theta\in\Theta}P_{N-1}(\theta)}
\exp(-\eta\Gamma_N^\theta(\gamma_N))\right)\,,
$$
and the usual analysis of the AA gives us the bound.
\end{proof}

Let us outline briefly how the construction
of Theorem~\ref{thm:AAfixed}
can be applied to non-mixable games
under the modified second-guessing protocol
(where experts are defined on $\partial\Sigma_\Lambda$).
Let the AA be $(c,\eta)$-realizable.
Now we are looking for $\gamma\in\partial\Sigma_\Lambda$
satisfying the following equation:
\begin{equation}\label{eq:AAfixednonmixable}
\gamma = V\left( F\left(-\frac{1}{\eta}\ln\left(\sum_{\theta\in\Theta}
\frac{P_{N-1}(\theta)}{\sum_{\theta\in\Theta}P_{N-1}(\theta)}
\exp(-\eta\Gamma_N^\theta(\gamma))\right)\right)\right)\,,
\end{equation}
where after $F$ we apply $V$, the mapping defined in
the proof of Lemma~\ref{lem:proj}.
Since $V$ is continuous and maps $\Sigma_\Lambda^\eta$
to $\partial\Sigma_\Lambda$,
we get a continuous mapping of 
$V(\Sigma_\Lambda^\eta)\subseteq\partial\Sigma_\Lambda$
into itself.
It remains to show that $V(\Sigma_\Lambda^\eta)$
has the fixed point property.
Similarly to the proof of Lemma~\ref{lem:proj},
consider the set 
$Z={\{g\in\Sigma_\Lambda^\eta\mid 
       \forall r\in[0,1)\:rg\notin\Sigma_\Lambda^\eta\}}$.
For any $g\in\Sigma_\Lambda^\eta$,
there exists a unique $r$ such that $rg\in Z$,
and the continuity of this mapping~$g\to rg$
follows in the same way 
as in the proof of Lemma~\ref{lem:proj};
thus $Z$ is a retract of $\Sigma_\Lambda^\eta$
and has the fixed point property.
Since $V(g)=V(rg)$ for any non-negative real $r$
such that $g$ and $rg$ belong to $\Sigma_\Lambda^\eta$,
we have $V(\Sigma_\Lambda^\eta)=V(Z)$.
The definition of $Z$ implies that $V$ is bijective on $Z$,
and again as in the proof of Lemma~\ref{lem:proj}
one can show that the inverse mapping 
$V^{-1}\colon V(Z)\to Z$
is continuous. Therefore $V(Z)$ has the fixed point property
as the homeomorphic image of~$Z$.

Let $\gamma_N\in V(\Sigma_\Lambda^\eta)$ 
be any solution of the equation~\eqref{eq:AAfixednonmixable}.
By the properties of $F$ and $V$,
we have
$$
\gamma_N \le -\frac{c}{\eta}\ln\left(\sum_{\theta\in\Theta}
\frac{P_{N-1}(\theta)}{\sum_{\theta\in\Theta}P_{N-1}(\theta)}
\exp(-\eta\Gamma_N^\theta(\gamma_N))\right)\,,
$$
and the usual AA bound follows.

\section{Predictions with Respect to Several Loss Functions}
\label{sec:multloss}

In this section, we illustrate the use
of the supermartingale technique
for another extension of Protocol~\ref{prot:PEA}:
a game with several loss functions
(for a more detailed discussion of this 
setting see~\cite{CV:2009}).
In contrast to the case of second-guessing
experts, it is not clear yet
whether the AA can help in this case.

Up to now a game was $(\Omega,\Lambda)$ where
$\Lambda$ was the set of admissible predictions,
common for Learner and Experts.
Here we return to the game specification
by a loss function on the decision space $\PPP(\Omega)$.
However, now each Expert $\theta$
has its own loss function $\lambda^\theta$.
So, the game is specified by
$(\Omega,\PPP(\Omega),\{\lambda^\theta\}_{\theta\in\Theta})$,
where 
$\lambda^\theta\colon\PPP(\Omega)\times\Omega\to[0,\infty]$
are proper loss functions.
The sets of predictions $\Lambda(\theta)$ and 
superpredictions $\Sigma_{\Lambda(\theta)}$
may be different for different experts $\theta$. 
The game follows Protocol~\ref{prot:PEEA}.

\begin{algorithm}[ht]
  \caption{Prediction with Expert Evaluators' Advice}
  \label{prot:PEEA}
  \begin{algorithmic}
    \STATE $L_0^{(\theta)}:=0$, for all $\theta\in\Theta$.
    \STATE $L_0^\theta:=0$, for all $\theta\in\Theta$.
    \FOR{$n=1,2,\dots$}
      \STATE All Experts $\theta\in\Theta$ announce 
                      $\pi_n^\theta\in\PPP(\Omega)$.
      \STATE Learner announces $\pi_n\in\PPP(\Omega)$.
      \STATE Reality announces $\omega_n\in\Omega$.
      \STATE $L_n^{(\theta)}:=L_{n-1}^{(\theta)}+\lambda^\theta(\pi_n,\omega_n)$, 
                                 for all $\theta\in\Theta$.
      \STATE $L_n^\theta:=L_{n-1}^\theta+\lambda^\theta(\pi_n^\theta,\omega_n)$, 
                                 for all $\theta\in\Theta$.
    \ENDFOR
  \end{algorithmic}
\end{algorithm}

There are two changes in Protocol~\ref{prot:PEEA}
compared to Protocol~\ref{prot:PEA}.
The accumulated loss~$L^\theta$ of each Expert~$\theta$ 
is calculated according to his own loss function~$\lambda^\theta$.
Learner does not have one accumulated loss anymore,
but the losses~$L^{(\theta)}$
of Learner are calculated separately for comparisons
with each Expert~$\theta$ and according to the loss function
of this Expert.

Now it does not make much sense to speak about the best expert: 
their performance is evaluated by different loss functions 
and thus the losses may have different scale.
What remains meaningful are bounds for every expert~$\theta$
of the form
$$
L^{(\theta)}_N\le c^\theta L_N^\theta + a^\theta\,,
$$
where $c^\theta$ and $a^\theta$ 
may be different for different experts $\theta\in\Theta$.

Informally speaking, Protocol~\ref{prot:PEEA} 
describes the following situation.
We have some practical task and 
a number of prediction algorithms (they will be our Experts).
Each of them minimizes some loss,
maybe different for different algorithms.
We do not know which algorithms fits our task best.
As usual in practice,
we do not have a loss that measures the quality of predictions
for our task; 
we only know that predictions must be close to the real outcomes.
A safe option in this case would be 
to predict in such a way that 
our prediction are not bad compared to predictions of 
any of the algorithms even if the quality is evaluated 
by the loss function ascribed to this algorithm.

The DFA can be adapted to Protocol~\ref{prot:PEEA}
straightforwardly.

\begin{theorem}\label{thm:multisuperbound}
Suppose that for each $\theta\in\Theta$,
there exist reals $c^\theta\ge 1$ and $\eta^\theta>0$
such that the functions
$$
\exp\left(\eta^\theta\biggl(\frac{\lambda^\theta(\pi,\omega)}{c^\theta}-
       g(\omega)\biggr)\right)
$$ 
(they are direct analogs of $q_g$ defined by~\eqref{eq:defsuperterm})
are forecast-continuous and have the supermartingale property 
for all $g\in\Sigma_{\Lambda^\theta}$.
Then for any initial distribution
$P_0\in\PPP(\Theta)$ 
there is Learner's strategy
(in fact, the DFA applied to $Q^{P_0}$
defined by~\eqref{eq:defmultisuper})
guaranteeing that,
at each step $N$ and for all experts $\theta$, it holds
$$
 L^{(\theta)}_N\le c^\theta L_N^\theta + 
    \frac{c^\theta}{\eta^\theta}\ln\frac{1}{P_0(\theta)}\,.
$$
\end{theorem}
\begin{proof}
Similarly to the proofs of Theorems~\ref{thm:superbound}
and~\ref{thm:supersecond},
we can construct the supermartingale $Q^{P_0}$:
\begin{multline}\label{eq:defmultisuper}
Q^{P_0}(\{\pi_1^\theta\}_{\theta\in\Theta},\pi_1,\omega_1,\ldots,
       \{\pi_N^\theta\}_{\theta\in\Theta},\pi_N,\omega_N)
=\\
\sum_{\theta\in\Theta}
P_0(\theta)\prod_{n=1}^N
\exp\left(\eta^\theta\biggl(\frac{\lambda^\theta(\pi_n,\omega_n)}{c^\theta}-
       \lambda^\theta(\pi_n^\theta,\omega_n)\biggr)\right)
\end{multline}
and choose $\pi_N$ satisfying~\eqref{eq:DFrealiz}
with the help of Lemma~\ref{lem:LevinSuper}.
The loss bound follows in the same way as in~Theorem~\ref{thm:superbound}.
\end{proof}

Protocol~\ref{prot:PEEA} can handle also the following task.
We have several experts and several candidates for the loss function,
and a priori some experts may perform well for two or more
of the loss functions.
In this case, it is natural to require that Learner's loss
is small with respect to every expert
and with respect to every loss function.
A simple trick reduces the task to Protocol~\ref{prot:PEEA}:
for each original expert (supplying us with a prediction),
we consider several new experts
who announce the same prediction
but use different loss functions.
If our predictions are good in the game with these new experts
then our predictions are good in the original game
with respect to any of the loss functions.

For example, assume that we want to 
compete with $K$ experts according to the 
logarithmic loss function and square loss function
in the game with outcomes $\{0,1\}$.
Lemmas~\ref{lem:supermartingale-log} 
and~\ref{lem:supermartingale-quadratic}
imply that the following function
is a forecast-continuous supermartingale:
\begin{multline*}
\frac{1}{2K}\sum_{k=1}^K\exp\left(\sum_{n=1}^N
\bigl(-\ln\pi_n(\omega_n)+\ln\pi_n^k(\omega_n)\bigr)
\right)\\
+
\frac{1}{2K}\sum_{k=1}^K\exp\left(2\sum_{n=1}^N
\bigl((\omega_n-\pi_n(1))^2 - (\omega_n-\pi_n^k(1))^2\bigr)
\right)\,,
\end{multline*}
where $\pi_n^k$ is the prediction of Expert $k$
and $\pi_n$ is the prediction of Learner.
Choosing $\pi_n$ according to Lemma~\ref{lem:LevinSuper},
we can achieve that the regret term
with respect to the logarithmic loss function is 
bounded by $\ln(2K)<\ln K + 0.7$,
and the regret 
with respect to the square loss function is 
bounded by $0.5\ln(2K)< 0.5\ln K + 0.4$~---
practically the same as the regrets against $K$ experts
that are achievable when we compete with respect to
only one of the loss functions.

\subsection*{Acknowledgements}
This work was partly supported by EPSRC grant
EP/F002998/1.
Discussions with Alex Gammerman, Glenn Shafer,
and Alexander Shen, and 
detailed comments of the anonymous referees
for the conference version~\cite{CKZV:2008} and 
for a journal submission have helped us improve the paper.

\appendix

\section*{Appendix}

\renewcommand{\proofname}{Proof of Lemma~\ref{lem:LevinProperty}}
\begin{proof}

Given the function $q$,
let us define the following function 
$\phi$ on ${\PPP(\Omega)\times\PPP(\Omega)}$:
$$
 \phi(\pi',\pi) = \EXP_{\pi'} q(\pi,\cdot)\,.
$$

For each fixed $\pi'$, the function $\phi(\pi',\cdot)$
is continuous, since $q$ is continuous.
For each fixed $\pi$, the function $\phi(\cdot,\pi)$
is linear, and thus concave.
Note also that $\PPP(\Omega)$ is a convex compact set.
Therefore, $\phi$ satisfies the conditions of 
Ky Fan's minimax theorem (see e.\,g.~\cite[Theorem 11.4]{agar:2001}),
and thus
there exists $\tilde\pi\in\PPP(\Omega)$
such that for any $\pi'\in\PPP(\Omega)$
it holds that 
\begin{equation}\label{eq:KyFan}
\EXP_{\pi'} q(\tilde\pi,\cdot)
=\phi(\pi',\tilde\pi)
\le \sup_{\pi\in\PPP(\Omega)} \phi(\pi,\pi)
= \sup_{\pi\in\PPP(\Omega)}\EXP_{\pi} q(\pi,\cdot)\le C\,.
\end{equation}

It is easy to see that $\tilde\pi$ has the property
that the lemma must guarantee:
$q(\tilde\pi,\omega)\le C$ for all $\omega\in\Omega$.
Indeed, if we substitute the distribution
$\delta_{\omega}$ (which is concentrated on $\omega$)
for $\pi'$ in~\eqref{eq:KyFan},
the left-hand side will be just
$q(\tilde\pi,\omega)$.
\end{proof}

Lemma~\ref{lem:LevinProperty} is a very important statement
in our supermartingale framework,
so let us outline an alternative proof for it
(for details see~\cite[Theorem 6]{gacs:2005},
\cite[Theorem~16.1]{GacsNotes}
or~\cite[Theorem 1]{Vovk:2007a}).
Consider the sets 
$F_\omega=\{\pi\mid q(\pi,\omega)\le C\}$.
These sets are closed and for any 
$\Omega_0\subseteq\Omega$
the union $\cup_{\omega\in\Omega_0} F_\omega$
contains all the measures concentrated on $\Omega_0$.
Then all $F_\omega$ has a non-empty intersection
by Sperner's lemma.

\begin{lemma}\label{lem:LevinEpsilon}
Let a function 
$q\colon\PPP^\circ(\Omega)\times\Omega\to\mathbb{R}$
be non-negative
and forecast-continuous on $\PPP^\circ(\Omega)$.
Suppose that for any $\pi\in\PPP^\circ(\Omega)$ it holds that
$$
 \EXP_\pi q(\pi,\cdot) \le C\,,
$$
where $C\in[0,\infty)$ is some constant.
Then it holds that
$$
\exists\pi\in\PPP^\circ(\Omega)\,
\forall\omega\in\Omega\quad
q(\pi,\omega)\le(1+\epsilon)C\,.
$$
\end{lemma}
\renewcommand{\proofname}{Proof}
\begin{proof}
Let $\delta\in(0,1)$ be a constant to be chosen later.

Let $\PPP^\delta(\Omega)=
{\{\pi\in\PPP(\Omega)\mid\forall\omega\in\Omega\;\pi(\omega)\ge\delta\}}$.
This set is a non-empty convex compact subset of $\PPP^\circ(\Omega)$.
Repeating the construction from the proof of Lemma~\ref{lem:LevinProperty}
and applying Ky Fan's theorem for the function on $\PPP^\delta(\Omega)$,
we get that
there exists $\tilde\pi\in\PPP^\delta(\Omega)$
such that for any $\pi'\in\PPP^\delta(\Omega)$
it holds that $\EXP_{\pi'} q(\tilde\pi,\cdot)\le C$.

For each $\omega_0$,
consider the distribution $\pi_{\delta,\omega_0}$ such that
$\pi_{\delta,\omega_0}(\omega)=\delta$ for $\omega\ne\omega_0$
and 
$\pi_{\delta,\omega_0}(\omega_0)={1-\delta(\lvert\Omega\rvert-1)}$.
Substituting $\pi_{\delta,\omega_0}$ for $\pi'$,
we get
$$
(1-\delta(\lvert\Omega\rvert-1))q(\tilde\pi,\omega_0)
+\delta\sum_{\omega\ne\omega_0}q(\tilde\pi,\omega)\le C\,.
$$
Since $q(\tilde\pi,\omega)\ge 0$ (the supermartingale $S$ is non-negative), 
the last inequality implies that
${(1-\delta(\lvert\Omega\rvert-1))q(\tilde\pi,\omega_0)}\le C$.
It remains to note that we can choose $\delta$ so small that
${1/(1-\delta(\lvert\Omega\rvert-1))}\le 1+\epsilon$.
\end{proof}

\renewcommand{\proofname}{Proof of Lemma~\ref{lem:LevinPropertyOpen}}
\begin{proof}
According to Lemma~\ref{lem:LevinEpsilon},
we can find $\pi_k\in\PPP^\circ(\Omega)$ such that 
$$
\forall\omega\in\Omega\quad
q(\pi_k,\omega)\le\left(1+\frac{1}{k}\right)C\,.
$$
Since $\PPP(\Omega)$ is compact,
there exists a strictly increasing index sequence 
$k(j)$, $j\in\mathbb{N}$, such that
the sequence $\pi_{k(j)}$ 
converges to some $\pi\in\PPP(\Omega)$.

The points $g_j=q(\pi_{k(j)},\cdot)$ 
belong to a compact set $[0,2C]^\Omega$.
Hence there exists
a strictly increasing index sequence 
$j(i)$, $i\in\mathbb{N}$, such that
the sequence $g_{j(i)}$ converges to some $g_0$.
For every $\omega\in\Omega$,
we have $g_j(\omega)=q(\pi_{k(j)},\omega)\le (1+1/k(j))C$,
therefore 
$$
g_0(\omega)=\lim_i g_{j(i)}(\omega)\le
\lim_i \left(1+\frac{1}{k(j(i))}\right)C=C\,.
$$

It remains to set $\pi^{(i)}=\pi_{k(j(i))}$
and note that $q(\pi^{(i)},\omega)=g_{j(i)}$.
\end{proof}

\renewcommand{\proofname}{Proof of Lemma~\ref{lem:proj}}
\begin{proof}
Let $\mathbf{0}\colon\Omega\to[0,\infty]$ be the constant
zero function 
(that is, ${\mathbf{0}(\omega)=0}$ for all $\omega\in\Omega$).
If $\mathbf{0}\in\Sigma_\Lambda$
then $\mathbf{0}\in\partial\Sigma_\Lambda$
and we can let $V(g)=\mathbf{0}$ for any $g$.

Assume that $\mathbf{0}\notin\Sigma_\Lambda$.
Let $V(g)=R(g)g$, where 
$R\colon\Sigma_\Lambda^\eta\to(0,c]$
is defined by the following rule:
$R(g)=\min\{r\in(0,c]\mid r g\in\Sigma_\Lambda\}$
for any $g\in\Sigma_\Lambda^\eta$.

Since the AA is $(c,\eta)$-realizable,
it holds that $c\Sigma_\Lambda^\eta\subseteq\Sigma_\Lambda$,
that is, $cg\in\Sigma_\Lambda$ for any $g\in\Sigma_\Lambda^\eta$.
The minimum is attained since $\Sigma_\Lambda$
is compact (by Assumption~\ref{assump1}).
Thus $R(g)$ is well defined.
It is obvious from the definition that 
$V(g)=R(g)g$ belongs to the boundary $\partial\Sigma_\Lambda$
of $\Sigma_\Lambda$ for all $g\in\Sigma_\Lambda^\eta$.

It remains to check that $V(g)=R(g)g$ is continuous in $g$.
We prove that $R$ is continuous,
namely, we take any $g_i\to g_0$
and for any infinite subsequence $\{g_{i_k}\}$,
we show that if $R(g_{i_k})$ converges
then $\lim_{k}R(g_{i_k})=R(g_0)$.
If $R(g_{i_k})$ converges then
$R(g_{i_k})g_{i_k}$ converges, 
and
$\lim_k R(g_{i_k})g_{i_k}=(\lim_k R(g_{i_k}))g_0\in\Sigma_\Lambda$
since $\Sigma_\Lambda$ is compact.
Therefore $R(g_0)\le\lim_k R(g_{i_k})$.
For the other inequality,
consider 
$R(g'|g)=\min\{r\in(0,\infty)\mid r g'\ge V(g)\}$
for $g,g'\in\Sigma_\Lambda^\eta$ such that
if $g(\omega)\ne 0$ for some $\omega\in\Omega$
then $g'(\omega)\ne 0$ too.
Clearly, the function $R(g'|g)$ is continuous in $g'$
for any fixed $g$
(note that
$R(g'|g)=\max_{\omega\colon g(\omega)\ne 0} V(g)(\omega)/g'(\omega)$)
and $R(g|g)=R(g)$.
Since $rg'\ge V(g)\in\Sigma_\Lambda$ 
implies $rg'\in\Sigma_\Lambda$,
we have $R(g'|g)\ge R(g')$.
In particular, 
$R(g_{i_k}|g_0)\ge R(g_{i_k})$
(assuming $k$ large enough so that
$g(\omega)\ne 0$ implies $g_{i_k}(\omega)\ne 0$)
and $R(g_0)=\lim_k R(g_{i_k}|g_0)\ge \lim_k R(g_{i_k})$.
\end{proof}

\renewcommand{\proofname}{Proof of Lemma~\ref{lem:ProperUnique}}
\begin{proof}
Assume that 
$\lambda_1(\pi,\omega_0)\ne\lambda_2(\pi,\omega_0)$
and $\pi(\omega_0)>0$
for some $\omega_0\in\Omega$.

Since $\lambda_1(\pi,\cdot)$ and $\lambda_2(\pi,\cdot)$
belong to $\Sigma_\Lambda^\eta$,
the point  
$$
g= -\frac{1}{\eta}
  \ln                                                   
  \frac{\e^{-\eta \lambda_1(\pi,\cdot)}
       +\e^{-\eta \lambda_2(\pi,\cdot)}}{2}
$$
also belongs to $\Sigma_\Lambda^\eta$
by the definition of $\Sigma_\Lambda^\eta$.

For any reals $x,y$, we have 
$(\e^x+\e^y)/2\ge \e^{(x+y)/2}$,
and the inequality is strict if $x\ne y$.
Therefore, 
$g(\omega)\le(\lambda_1(\pi,\omega)+\lambda_2(\pi,\omega))/2$
for all $\omega\in\Omega$
and                                                          
$g(\omega_0)<(\lambda_1(\pi,\omega_0)+\lambda_2(\pi,\omega_0))/2$.
Multiplying these inequalities by $\pi(\omega)$
and summing over all $\omega\in\Omega$,
we get 
$$
\EXP_\pi g  
 < 
  \frac{1}{2}
   \bigl(  \EXP_\pi \lambda_1(\pi,\cdot)
            + \EXP_\pi \lambda_2(\pi,\cdot)
   \bigr)
$$
(recall that $\pi(\omega_0)>0$).
Since $\lambda_1$ and $\lambda_2$ are proper 
with respect to~$\Sigma_\Lambda^\eta$,
we have $\EXP_\pi \lambda_1(\pi,\cdot)\le \EXP_\pi g$
and $\EXP_\pi \lambda_2(\pi,\cdot)\le \EXP_\pi g$.
Hence we get a contradiction ${\EXP_\pi g < \EXP_\pi g}$.
\end{proof}

For a convex function 
$U\colon\mathbb{R}^\Omega\to[-\infty,\infty]$,
a \emph{subgradient} at point $x\in\mathbb{R}^\Omega$
is a point $x^\ast\in\mathbb{R}^\Omega$ such that
$$
 \forall z\in\mathbb{R}^\Omega\:\:
 U(z)\ge U(x) + \langle x^\ast,z-x\rangle\,.
$$

\begin{lemma}\label{lem:convex}
Suppose that $Y$ is a non-empty closed convex subset of $[0,\infty)^\Omega$.
Let ${U\colon\mathbb{R}^\Omega\to(-\infty,\infty]}$ be the function
$$
 U(x) = - \inf_{y\in Y} \langle x,y\rangle\,,
$$
where $\langle x,y\rangle = \sum_{\omega\in\Omega} x(\omega)y(\omega)$
is the scalar product in $\mathbb{R}^\Omega$.
Then $U(x)$ is a convex function,
and for any $\pi\in\PPP(\Omega)$,
it holds that ${U(\pi)<\infty}$,
and $\pi^\ast$ is a subgradient of $U$ at the point $\pi$
if and only if 
$-\pi^\ast\in Y$ and ${\langle\pi,-\pi^\ast\rangle = -U(\pi)}$.
\end{lemma}

\renewcommand{\proofname}{Proof}
\begin{proof}
Since $Y$ is not empty, 
the infimum is finite, 
and therefore $U(x)> -\infty$ for all $x$.

For any $\alpha\in[0,1]$ and any $x_1,x_2\in\mathbb{R}^\Omega$,
we have 
$U(\alpha x_1+(1-\alpha)x_2)=
-\inf_{y\in Y}(\alpha\langle x_1,y\rangle+(1-\alpha)\langle x_2,y\rangle)
\le 
-\inf_{y\in Y}\alpha\langle x_1,y\rangle
-\inf_{y\in Y}(1-\alpha)\langle x_2,y\rangle
=\alpha U(x_1)+(1-\alpha)U(x_2)$,
thus $U$ is convex.

Let us fix some $\pi\in\PPP(\Omega)$.
Then $\langle \pi,y\rangle \ge 0$ for all $y\in Y$,
and $U(\pi)\le 0<\infty$.

Let $-\pi^\ast\in Y$ and ${\langle\pi,-\pi^\ast\rangle = -U(\pi)}$.
Then 
$U(\pi) + \langle \pi^\ast,z-\pi\rangle
= - \langle -\pi^\ast,z\rangle \le
- \inf_{y\in Y} \langle z,y\rangle = U(z)$
for any $z$,
thus $\pi^\ast$ is a subgradient of $U$ at $\pi$.

Let $\pi^\ast$ be any subgradient of $U$ at $\pi$.
Assume that $-\pi^\ast\notin Y$.
Then $-\pi^\ast$ and $Y$ can be strongly separated by Corollary~11.4.2 
in~\cite{Rock:1970}, and Theorem~11.1(c) there implies
that there exists $z\in\mathbb{R}^\Omega$ such that
$\inf_{y\in Y} \langle y,z\rangle > \langle -\pi^\ast,z\rangle$.
Let us choose $\delta>0$ such that 
$$
\inf_{y\in Y} \langle y,z\rangle > \delta + \langle -\pi^\ast,z\rangle\,,
$$
and then choose $y_0\in Y$ such that
$$
\langle\pi+z,y_0\rangle < \inf_{y\in Y} \langle\pi+z,y\rangle + \delta\,.
$$
From the definition of the subgradient, we get
$U(\pi+z)\ge U(\pi) + \langle \pi^\ast,z\rangle$,
and thus
$$
\langle\pi+z,y_0\rangle - \delta < 
\inf_{y\in Y}\langle\pi+z,y\rangle\le 
\inf_{y\in Y}\langle\pi,y\rangle + \langle -\pi^\ast,z\rangle
\le \langle\pi,y_0\rangle + \langle -\pi^\ast,z\rangle\,.
$$
So, $\langle z,y_0\rangle < \delta + \langle -\pi^\ast,z\rangle$,
which contradicts the choice of $\delta$.
This means that $-\pi^\ast\in Y$.

It remains to note that the definition of the subgradient 
implies $U(0)\ge U(\pi) + \langle \pi^\ast,0-\pi\rangle$,
and since $U(0)=0$, 
we get 
$\inf_{y\in Y} \langle y,\pi\rangle =
-U(\pi)\ge \langle -\pi^\ast,\pi\rangle$.
\end{proof}

\renewcommand{\proofname}{Proof of Lemma~\ref{lem:ProperNonzero}}
\begin{proof}
By Assumption~\ref{assump2}, 
there exists a finite point $g_{\mathrm{fin}}$ in 
$\Sigma_\Lambda^\eta\cap[0,\infty)^\Omega$,
where $\EXP_\pi g_{\mathrm{fin}}$ is finite for any $\pi$.
By Assumption~\ref{assump1}, $\Sigma_\Lambda^\eta$
is compact, and therefore the minimum is attained for all 
$\pi\in\mathbb{R}^\Omega$.
Thus $H$ is well defined.
Note also that $H(\pi)\ge 0$ for $\pi\in\PPP(\Omega)$
and $H(\pi)=-\infty$ if $\pi(\omega)<0$ for some $\omega\in\Omega$.

Now let us show that 
$$
H(\pi) = 
  \inf_{g\in\Sigma_\Lambda^\eta\cap[0,\infty)^\Omega} \EXP_\pi g\,.
$$
Again by Assumption~\ref{assump2}, 
the infimum is taken over a non-empty set.
If $\pi(\omega)<0$ for some $\omega\in\Omega$
then $H(\pi)=-\infty$ and the infimum is equal to $-\infty$ as 
well.
Thus we need to consider only the case when
$\pi(\omega)\ge 0$ for all $\omega\in\Omega$
and the minimum in the definition of $H$ is attained
at a point $g$ such that $g(\omega)=\infty$ for some values of $\omega$.
Note that for these $\omega$ we have $\pi(\omega)=0$,
since $H(\pi)< \infty$.
Choose a sequence $g_n\in\Sigma_\Lambda^\eta\cap[0,\infty)^\Omega$
that converges to $g$ 
(for example, consider the segment between 
the points $\e^{-\eta g}$ and $\e^{-\eta g_{\mathrm{fin}}}$, 
and take a sequence $\e^{-\eta g_n}$ along this segment).
Since $g_n(\omega)$ and $g(\omega)$ are finite 
for non-zero $\pi(\omega)$, we get $\EXP_\pi g_n\to\EXP_\pi g=H(\pi)$,
and thus the infimum is not greater than $H(\pi)$.

Now we can apply Lemma~\ref{lem:convex} with
$Y=\Sigma_\Lambda^\eta\cap[0,\infty)^\Omega$
and $U(\pi)=-H(\pi)$.
It implies that for any $\pi\in\PPP(\Omega)$,
the set of subgradients of $U$ at $\pi$
is the set of points where the infimum of $\EXP_\pi g$
over $g\in\Sigma_\Lambda^\eta\cap[0,\infty)^\Omega$
is attained.
If $\pi\in\PPP^\circ(\Omega)$, the infimum is attained indeed,
and it is unique by Lemma~\ref{lem:ProperUnique}.
By Theorem~25.1 in~\cite{Rock:1970},
the function $H$ is differentiable at $\pi$,
and the point
$\lambda(\pi,\cdot)=\arg\min_{g\in\Sigma_\Lambda^\eta}\EXP_\pi g$ 
is the gradient of $H$.
Thus $\mathcal{H}\supseteq\PPP^\circ(\Omega)$.

On the other hand, if $H$ is differentiable, the set of subgradients
consists of one element only, the gradient.
Theorem~25.5 in~\cite{Rock:1970} implies that 
the gradient mapping 
$\pi\mapsto\lambda(\pi,\cdot)$
is continuous on $\mathcal{H}$.
\end{proof}

\renewcommand{\proofname}{Proof of Lemma~\ref{lem:ProperLimit}}
\begin{proof}
Due to Assumption~\ref{assump1}, $\Omin\Sigma_\Lambda^\eta$
is compact and therefore contains all its limit points,
that is, $\gamma\in\Omin\Sigma_\Lambda^\eta$.

Let $\EXP^{+}_{\pi'}g$ be a shorthand 
for $\sum_{\omega\in\Omega,\:\pi(\omega)\ne0}\pi'(\omega)g(\omega)$
for any $g\in[0,\infty]^\Omega$ and $\pi'\in\PPP(\Omega)$.
By definition,
$\EXP_\pi\gamma = \EXP^{+}_\pi\gamma$.

Note first that $\EXP_{\pi_i}g$ converges to $\EXP_{\pi}g$
for any finite $g\in[0,\infty)^\Omega$.

Note also that 
$\EXP^{+}_{\pi_i}\gamma_i$ 
converges to 
$\EXP^{+}_{\pi}\gamma$.
Indeed, 
$\EXP^{+}_{\pi_i}\gamma_i\le \EXP_{\pi_i}\gamma_i
\le \EXP_{\pi_i}g_{\mathrm{fin}}
\le \sum_{\omega\in\Omega}g_{\mathrm{fin}}(\omega)
<\infty$,
where $g_{\mathrm{fin}}\in\Sigma_\Lambda^\eta\cap[0,\infty)^\Omega$
exists by Assumption~\ref{assump2}.
If $\pi(\omega)\ne 0$ then $\pi_i(\omega)$ is separated from~$0$
for sufficiently large~$i$, therefore $\gamma_i(\omega)$ are bounded,
and their limit $\gamma(\omega)$ is finite.
And for finite limits $\gamma$ and $\pi$, the convergence
is trivial.

Fix any $g_0\in[0,\infty)^\Omega$ and any $\epsilon>0$. 
For sufficiently large~$i$, we have 
$\EXP^{+}_{\pi_i}\gamma_i\ge \EXP^{+}_{\pi}\gamma - \epsilon$
and 
$\EXP_{\pi_i}g_0 \le \EXP_{\pi}g_0 + \epsilon$.
Taking into account that 
$\EXP^{+}_{\pi_i}\gamma_i\le \EXP_{\pi_i}\gamma_i$
and 
$\EXP_{\pi_i}\gamma_i=\min_{g\in\Sigma_\Lambda^\eta} \EXP_{\pi_i}g
\le \EXP_{\pi_i}g_0$,
we get 
$\EXP_\pi\gamma\le \EXP_{\pi}g_0 + 2\epsilon$.
Since $\epsilon$ and $g_0$ are arbitrary,
we have 
$$
\EXP_\pi\gamma\le 
 \inf_{g\in\Sigma_\Lambda^\eta\cap[0,\infty)^\Omega}\EXP_{\pi}g\,,
$$
and the last infimum can be replaced by 
$\min_{g\in\Sigma_\Lambda^\eta}$ 
as shown in the proof of Lemma~\ref{lem:ProperNonzero}.
\end{proof}

\renewcommand{\proofname}{Proof of Lemma~\ref{lem:AAcont}}
\begin{proof}
We construct a continuous mapping 
$F\colon\Sigma_\Lambda^\eta\to\Omin\Sigma_\Lambda^\eta$ 
as a composition of mappings $F_\omega$ for all $\omega\in\Omega$.
Each $F_\omega$ when applied to $g\in\Sigma_\Lambda^\eta$
preserves the values of $g(o)$ for $o\ne\omega$
and decreases as far as possible the value $g(\omega)$ 
so that the result is still in $\Sigma_\Lambda^\eta$.
Formally, $F_\omega(g)=g'$ such that
$g'(o)=g(o)$ for $o\ne\omega$ and
$g'(\omega)=
\min\{\,\gamma(\omega)\mid \gamma\in\Sigma_\Lambda^\eta,\:
     \forall o\ne\omega\: \gamma(o)=g(o)\:\}$.

Let us show that each $F_\omega$ is continuous. 
It suffices to show that 
$F_\omega(g)(\omega)$ depends continuously on $g$,
since the other coordinates do not change.
We will show that $F_\omega(g)(\omega)$ is convex in $g$,
continuity follows (see, e.\,g.~\cite{Rock:1970}).
Indeed, take any $t\in[0,1]$, and $g_1,g_2\in\Sigma_\Lambda^\eta$.
Since $\Sigma_\Lambda^\eta$ is convex, 
then $tg_1+(1-t)g_2\in\Sigma_\Lambda^\eta$ and 
$tF_\omega(g_1)+(1-t)F_\omega(g_2)\in\Sigma_\Lambda^\eta$.
The latter point has all the coordinates $o\ne\omega$
the same as the former. 
Thus, by definition of $F_\omega$,
we get
$F_\omega(tg_1+(1-t)g_2)(\omega)\le(tF_\omega(g_1)+(1-t)F_\omega(g_2))(\omega)=
tF_\omega(g_1)(\omega)+(1-t)F_\omega(g_2)(\omega)$, 
which was to be shown.

All $F_\omega$ do not increase the coordinates.
Since the set $\Sigma_\Lambda^\eta$ contains any point $g$
with all its majorants,
$F_\omega(g_1)=g_1$ implies that $F_\omega(g_2)=g_2$
for any $g_2$ obtained from $g_1$ by applying any $F_{\omega'}$.
Therefore, 
the image of a composition of $F_\omega$ over all $\omega\in\Omega$
is included in $\Omin\Sigma_\Lambda^\eta$.
\end{proof}

\end{document}